\documentclass[utf8]{frontiersSCNS} 

\usepackage{url,hyperref,lineno,microtype,subcaption}
\usepackage[onehalfspacing]{setspace}

\def\keyFont{\fontsize{8}{11}\helveticabold }
\def\firstAuthorLast{Srinivasan {et~al.}} 
\def\Authors{Gopalakrishnan Srinivasan\,$^{1,*}$ and Kaushik Roy\,$^{1}$}
\def\Address{$^{1}$Department of ECE, Purdue University, West Lafayette, IN, USA}
\def\corrAuthor{465 Northwestern Ave, West Lafayette, IN 47907}

\def\corrEmail{srinivg@purdue.edu}

\begin{document}
\onecolumn
\firstpage{1}

\title[ReStoCNet]{ReStoCNet: Residual Stochastic Binary Convolutional Spiking Neural Network for Memory-Efficient Neuromorphic Computing} 

\author[\firstAuthorLast ]{\Authors} 
\address{} 
\correspondance{} 

\extraAuth{}

\maketitle

\begin{abstract}
In this work, we propose ReStoCNet, a residual stochastic multilayer convolutional Spiking Neural Network (SNN) composed of binary kernels, to reduce the synaptic memory footprint and enhance the computational efficiency of SNNs for complex pattern recognition tasks. ReStoCNet consists of an input layer followed by stacked convolutional layers for hierarchical input feature extraction, pooling layers for dimensionality reduction, and fully-connected layer for inference. In addition, we introduce residual connections between the stacked convolutional layers to improve the hierarchical feature learning capability of deep SNNs. We propose Spike Timing Dependent Plasticity (STDP) based probabilistic learning algorithm, referred to as Hybrid-STDP (HB-STDP), incorporating Hebbian and anti-Hebbian learning mechanisms, to train the binary kernels forming ReStoCNet in a layer-wise unsupervised manner. We demonstrate the efficacy of ReStoCNet and the presented HB-STDP based unsupervised training methodology on the MNIST and CIFAR-10 datasets. We show that residual connections enable the deeper convolutional layers to self-learn useful high-level input features and mitigate the accuracy loss observed in deep SNNs devoid of residual connections. The proposed ReStoCNet offers $>$20$\times$ kernel memory compression compared to full-precision (32-bit) SNN while yielding high enough classification accuracy on the chosen pattern recognition tasks.
\tiny
 \keyFont{ \section{Keywords:} Convolutional SNN, Spiking ResNet, Binary Kernels, Probabilistic STDP, Unsupervised Feature Learning}
\end{abstract}

\section{Introduction}
The proliferation in real-time content generated by the ubiquitous battery-powered edge devices necessitates a paradigm shift in neural architectures to enable energy-efficient neuromorphic computing. Spiking Neural Networks (SNNs) offer a promising alternative towards realizing intelligent neuromorphic systems that require lower computational effort than the artificial neural networks. SNNs encode and communicate information in the form of sparse spiking events. The intrinsic sparse event-driven processing capability, which entails neuronal computations and synaptic weight updates only in the event of a spike fired by the constituting neurons, leads to improved energy efficiency in neuromorphic hardware implementations \citep{sengupta2019going}. Spike Timing Dependent Plasticity (STDP) \citep{bi1998synaptic} is a localized hardware-friendly plasticity mechanism used for unsupervised learning in SNNs. STDP-based learning rules \citep{song2000competitive} modify the weight of a synapse interconnecting a pair of input (pre) and output (post) neurons depending on the degree of correlation between the respective spike times. The spike timing information is encoded in the bit-precision of the synaptic weight. In an effort to reduce the synaptic memory footprint, \citet{suri2013bio}, \citet{querlioz2015bioinspired}, and \citet{srinivasan2016magnetic} proposed two-layer fully-connected SNN composed of binary synaptic weights. The fully-connected SNN learns complete input representations rather than distinctive features making up the input patterns. As a result, it requires large number of trainable parameters to attain competitive classification accuracy \citep{diehl2015unsupervised}, which negatively impacts the scalability of such shallow SNNs for complex pattern recognition tasks.

We propose deep \underline{Re}sidual \underline{Sto}chastic Binary \underline{C}onvolutional Spiking Neural \underline{Net}work, referred to as \textit{ReStoCNet}, as a scalable architecture to achieve improved classification accuracy with compressed synaptic memory. ReStoCNet consists of an input layer followed by stacked convolutional layers with Leaky-Integrate-and-Fire (LIF) spiking non-linearity \citep{dayan2001theoretical} for hierarchical input feature extraction, spatial pooling layers for dimensionality reduction, and one or more fully-connected layers for inference. We introduce residual or shortcut connections between the stacked convolutional layers, inspired by the organization of deep residual networks \citep{he2016deep}, in order to improve the representations learnt by the later convolutional layers. In addition, we enforce binary synaptic weights for the convolutional kernels during both training and inference. We propose STDP-based probabilistic learning rule, referred to as Hybrid-STDP (HB-STDP), incorporating Hebbian and anti-Hebbian learning mechanisms to train the binary kernels. Based on HB-STDP, a binary synaptic weight is probabilistically potentiated for small positive time difference between excitatory pre- and post-spikes, which is in agreement with the Hebbian learning theory \citep{hebb1949organization}. On the other hand, it is probabilistically depressed for large positive time difference (anti-Hebbian in nature) or small negative time difference (Hebbian in nature) between the respective spikes. The spike timing information is essentially encoded in the synaptic switching probability, which is held constant within the Hebbian potentiation, Hebbian depression, and anti-Hebbian depression windows, and is zero elsewhere. We note that \citet{suri2013bio} proposed an STDP-based learning rule employing constant switching probabilities, where the potentiation and depression windows extend over the entire STDP timing window. On the contrary, HB-STDP contains dead zone in the STDP timing window, where the switching probability is zero. We visually demonstrate the significance of dead zone for efficient feature learning using binary fully-connected SNN.

We present HB-STDP based layer-wise unsupervised training methodology for ReStoCNet, where we train the binary kernels interconnecting successive convolutional layers using HB-STDP. Once a given layer is trained, we forward propagate the spikes from the input through the trained layers and update the binary kernels of the following convolutional layer. After all the convolutional layers are trained, we feed the input dataset, estimate the spiking activations of the spatially pooled convolutional spike maps by accumulating the spikes at every time instant and decaying the resultant sum between successive spike timing instants, and pass them on to the fully-connected layer, trained using error backpropagation \citep{rumelhart1986learning}, for inference. We validate the efficacy of ReStoCNet and the HB-STDP based unsupervised training methodology on the MNIST \citep{lecun1998gradient} and CIFAR-10 datasets \citep{krizhevsky2009learning}. We show that residual connections enable the deeper convolutional layers to extract useful high-level input features and effectively mitigate the accuracy degradation observed in deep SNNs devoid of residual connections \citep{lee2018deep}. We note that \citet{masquelier2007unsupervised}, \citet{panda2016unsupervised}, \citet{lee2016training}, \citet{stromatias2017event}, \citet{srinivasan2018stdpbased}, \citet{tavanaei2018training}, \citet{kheradpisheh2018stdp}, \citet{ferre2018unsupervised}, \citet{thiele2018event}, \citet{lee2018deep, lee2018training}, and \citet{mozafari2018combining} have demonstrated convolutional SNNs composed of full-precision kernels. Recently, \citet{sengupta2019going} and \citet{hu2018spiking} presented residual SNNs, trained using error backpropagation with real-valued inputs and artificial ReLU neurons \citep{nair2010rectified}, which are mapped to spiking neurons post training for energy-efficient inference. To the best of our knowledge, ReStoCNet is the first demonstration of STDP-trained deep residual convolutional SNN composed of binary kernels for complex pattern recognition tasks. We believe that ReStoCNet, with event-driven computing capability and memory-efficient learning with binary kernels trained using hardware-friendly probabilistic-STDP learning rule, offers a promising alternative for energy-efficient neuromorphic computing in battery-powered edge devices. Overall, the key contributions of our work are:
\begin{enumerate}
\item We propose ReStoCNet, a deep residual convolutional SNN composed of binary kernels, for memory-efficient neuromorphic computing.
\item We present HB-STDP, an STDP-based probabilistic learning rule incorporating Hebbian and anti-Hebbian learning mechanisms, for training the binary kernels constituting ReStoCNet in a layer-wise unsupervised manner for hierarchical input feature extraction.
\item We validate the efficacy of ReStoCNet on the MNIST and CIFAR-10 datasets, and show that residual connections enable the deeper convolutional layers to learn useful high-level input features and mitigate the accuracy loss incurred by STDP-trained deep SNNs without residual connections. 
\end{enumerate}

\begin{figure}[!b]
\centering
\includegraphics[width=7.0in]{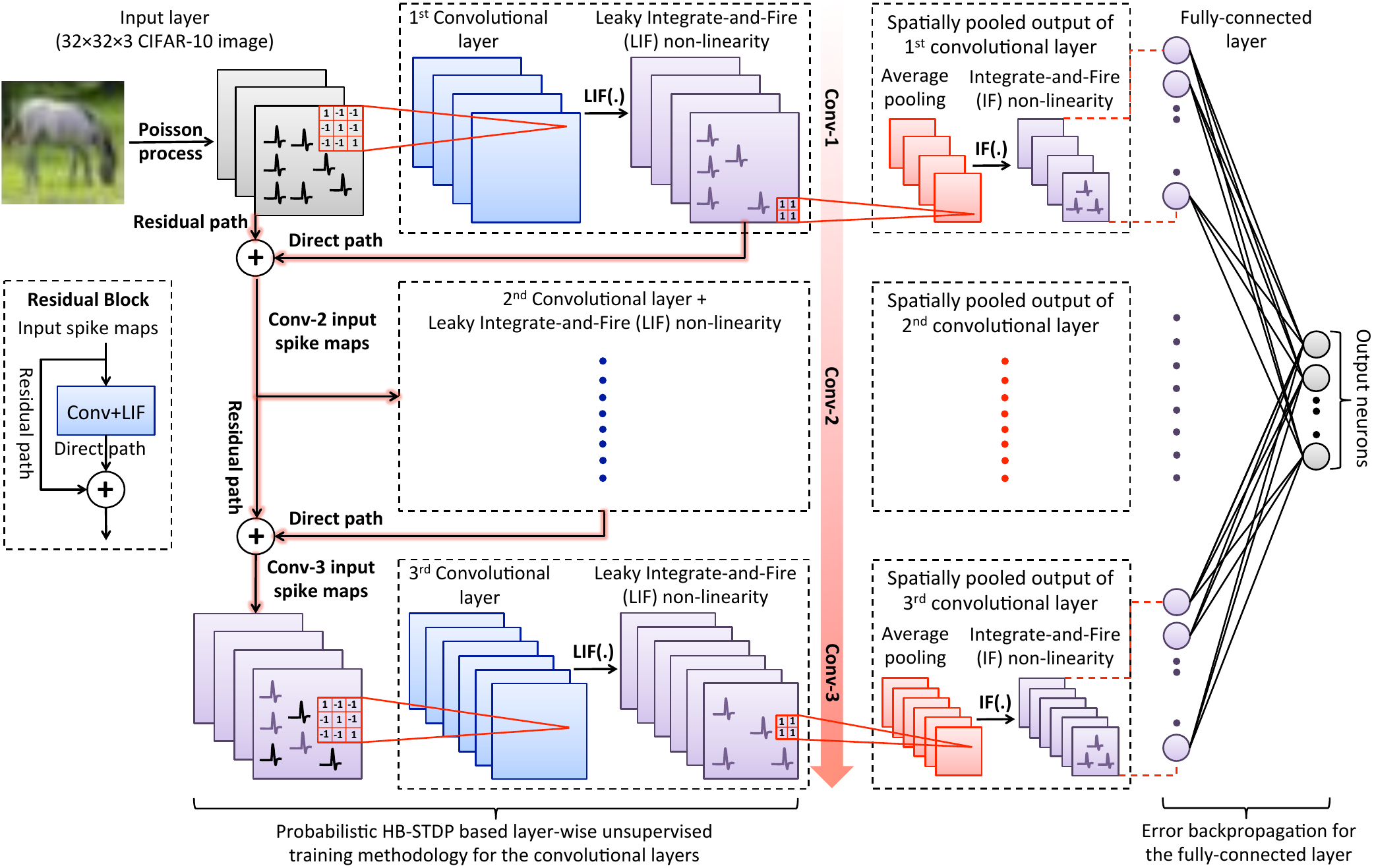}
\caption{Illustration of ReStoCNet consisting of an input layer followed by stacked convolutional layers with Leaky-Integrate-and-Fire (LIF) spiking non-linearity, which are interconnected via binary kernels. The deeper convolutional layers receive residual inputs that are summed up with direct inputs from the preceding convolutional layer as depicted in the inset. The binary kernels forming the convolutional layers are trained using probabilistic Hybrid-STDP (HB-STDP) based layer-wise unsupervised training methodology. After all the convolutional layers are trained, the respective spike maps are spatially pooled using average pooling with 2$\times$2 unit-weight kernels followed by Integrate-and-Fire (IF) spiking non-linearity to produce the pooled spike maps. The spike trains of the pooling layers are low-pass filtered to obtain their spiking activations over the time period for which the input is presented, which are fed to the fully-connected layer, trained using error backpropagation, for inference.}
\label{fig:ReStoCNet_arch}
\end{figure}

\section{Material and Methods}
\subsection{ReStoCNet: \underline{Re}sidual \underline{Sto}chastic Binary \underline{C}onvolutional Spiking Neural \underline{Net}work} \label{sec:ReStoCNet_arch}
ReStoCNet consists of an input layer followed by stacked convolutional layers for hierarchical input feature extraction, spatial pooling layers for dimensionality reduction, and one or more fully-connected layers for inference as illustrated in Fig. \ref{fig:ReStoCNet_arch}. The pixels in the input image maps are converted to Poisson spike trains firing at a rate proportional to the corresponding pixel intensities. At any given time, the input spike maps are convolved with the binary kernels, which are constrained to logic states $-$1 ($w_{low}$) and $+$1 ($w_{high}$), to produce the convolutional output maps. The convolutional outputs, referred to as post-synaptic currents, are fed to non-linear layer of Leaky-Integrate-and-Fire (LIF) spiking neurons \citep{dayan2001theoretical}. An LIF neuron integrates the post-synaptic current into its membrane potential, whose dynamics are described by
\begin{equation} \label{eq:LIF}
\tau_{mem}\frac{dV_{mem}}{dt} = -V_{mem} + I_{post}
\end{equation}
where $V_{mem}$ is the neuronal membrane potential, $\tau_{mem}$ is the membrane potential leak time constant, and $I_{post}$ is the post-synaptic current. The LIF neuron emits a spike when its membrane potential exceeds a definite firing threshold after which the membrane potential is reset to zero. Every convolutional output map yields a corresponding spike map based on the LIF spiking neuronal dynamics, which is directly fed to the following convolutional layer. In addition, we introduce residual connections feeding into the deeper convolutional layers, which is inspired by the architecture of deep residual networks \citep{he2016deep}. The second convolutional layer receives residual connections from the input layer while the third convolutional layer receives residual connections from the input and first convolutional layer as shown in Fig. \ref{fig:ReStoCNet_arch}. The residual connections feeding into a target convolutional layer perform identity mapping, i.e., the residual path spike maps are simply added to the direct path spike maps from the preceding convolutional layer and fed to the target convolutional layer. In the event of a mismatch in the number of spike maps (or channels) between the residual and direct paths, the spike maps in the residual path are replicated to be consistent with the number of channels in the direct path. Consider, for instance, the second convolutional layer that receives spike maps from the input layer via the residual path and the first convolutional layer via the direct path. Let us suppose that the input image pattern is stored in RGB colorspace. Consequently, each image pattern yields 3 input spike maps that needs to be summed up with the spike maps of the first convolutional layer, which typically contains more than 3 spike maps. Hence, the 3 input spike maps are replicated to match the number of spike maps in the first convolutional layer, summed up with the spike maps of the first convolutional layer, and fed to the second convolutional layer. Note that the summed spike maps from the residual and direct paths are constrained to unit magnitude to produce resultant spike maps feeding into the target convolutional layer. The binary kernels constituting the convolutional layers are trained using probabilistic Hybrid-STDP (HB-STDP) based layer-wise unsupervised training methodology. We find that the residual connections ensure rich and diverse inputs for deeper convolutional layers and enable them to self-learn useful high-level input features as shown in \autoref{sec:ReStoCNet_CIFAR}. The improved feature learning capability mitigates the accuracy loss incurred by stacked convolutional layers without residual connections as experimentally validated in \autoref{sec:ReStoCNet_CIFAR} and enhances the scalability of deep SNNs.

After all the convolutional layers are trained, we feed the input dataset and spatially pool the spike maps of the convolutional layers. Spatial pooling is the mechanism used to suitably combine the neighboring pixels of a convolutional feature map to reduce the map size (height and width) while retaining the salient features. Spatial pooling also renders the network invariant to slight translations in the input features \citep{jaderberg2015spatial}. We perform a class of spatial pooling operation known as average pooling with 2$\times$2 kernels composed of unit weights and stride length of 2 as detailed below. The spikes in every 2$\times$2 non-overlapping region of the convolutional maps are summed up and normalized by the kernel size (4 for a 2$\times$2 kernel) to produce the pooled output maps, which are then fed to a layer of Integrate-and-Fire (IF) spiking neurons to generate the pooled spike maps. An IF neuron integrates the input into its membrane potential and spikes if the membrane potential exceeds pre-specified threshold ($\theta_{pool}$) after which the membrane potential is reset. The IF neurons, in effect, fire based on the average spiking activity of the spatially pooled convolutional spike maps. We low-pass filter the spike trains of the pooled maps by integrating the spikes at every time instant and decaying the resultant sum between successive spike timing instants to estimate their spiking activations over the time period for which the input is presented. The spiking activations of the pooled maps pertaining to all the convolutional layers are fed to the fully-connected layer composed of ReLU neurons \citep{nair2010rectified} for inference. This ensures that the input features learnt independently by the convolutional layers in an unsupervised manner are combined optimally by the fully-connected layer to yield the best accuracy. We note that LIF neurons can instead be used in the fully-connected layer, which can be trained using spike-based backpropagation algorithms \citep{lee2016training, panda2016unsupervised, wu2018spatio, lee2018training, jin2018hybrid}. In this work, we use fully-connected layer of ReLU neurons trained with backpropagation algorithm commonly used for deep learning networks since we are primarily interested in evaluating the efficacy of the proposed probabilistic HB-STDP based unsupervised training methodology for the convolutional layers that is detailed in the following subsection.

\subsection{Hybrid-STDP (HB-STDP) for Binary Synaptic Weights} \label{sec:HB-STDP}
We propose STDP-based probabilistic learning rule, referred to as Hybrid-STDP (HB-STDP), integrating Hebbian and anti-Hebbian learning mechanisms to train the binary synaptic weights constituting an SNN. We present two versions of the HB-STDP learning rule, namely, excitatory HB-STDP (eHB-STDP) and inhibitory HB-STDP (iHB-STDP) to train the binary synaptic weights connecting excitatory and inhibitory pre-neurons, respectively, to excitatory post-neurons. An excitatory neuron is modeled as a neuron firing unit positive spikes while an inhibitory neuron fires unit negative spikes. Input image pixels with intensities ranging from 0 to 255 are mapped to excitatory pre-neurons firing unit positive spikes at a rate proportional to the respective pixel intensities. On the contrary, input images when pre-processed by normalizing the raw pixel intensities to zero mean and unit variance result in normalized images with positive and negative pixel intensities. The normalized pixels with negative intensities are mapped to inhibitory pre-neurons firing unit negative spikes. The normalized input maps containing excitatory and inhibitory pre-neurons offer richer spike-encoding of the image patterns, resulting in efficient STDP-based feature learning. We find that input normalization is critical for natural images like those from the CIFAR-10 dataset \citep{krizhevsky2009learning} that do not have clear separation between the region of interest and the background unlike digit patterns from the MNIST dataset \citep{lecun1998gradient}.

\begin{figure}[!t]
\centering
\includegraphics[width=7.0in]{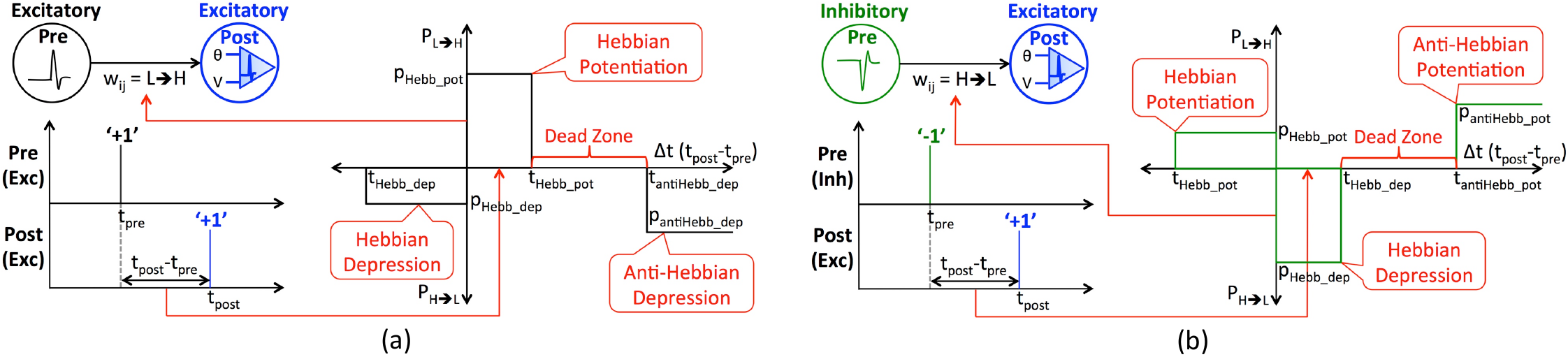}
\caption{(a) Illustration of eHB-STDP, an STDP-based probabilistic learning rule, incorporating Hebbian and anti-Hebbian learning mechanisms, for training the binary synaptic weights interconnecting excitatory pre- and post-neurons firing positive spikes. The synaptic weight is probabilistically potentiated for small positive time difference (Hebbian in nature) while it is probabilistically depressed for large positive (anti-Hebbian in nature) or small negative time difference (Hebbian in nature) between the pre- and post-spikes. The switching probability is held constant within the Hebbian potentiation, Hebbian depression, and anti-Hebbian depression windows, and is zero in the dead zone. (b) Illustration of iHB-STDP for binary synaptic weights connecting inhibitory pre-neurons firing negative spikes to excitatory post-neurons. The iHB-STDP dynamics are obtained by mirroring the eHB-STDP dynamics about the $\Delta t$ ($t_{post}-t_{pre}$) axis.}
\label{fig:HB_STDP}
\end{figure}
Binary synapses require a probabilistic learning rule to prevent rapid switching of the weights between the allowed levels, which could otherwise render the synapses memoryless. Both the proposed eHB-STDP and iHB-STDP learning rules map the time difference between a pair of pre- and post-spikes to the switching probability of the interconnecting binary synapse. We first detail the eHB-STDP learning rule for excitatory pre-neurons and subsequently discuss how the learning dynamics are adapted for inhibitory pre-neurons. According to eHB-STDP, if an excitatory pre-spike (at time instant, $t_{pre}$) triggers the post-neuron to fire (at time instant, $t_{post}$) and the difference between the respective spike times ($\Delta t = t_{post}-t_{pre}$) is smaller than a pre-specified time period ($t_{Hebb\_pot}$), we switch the synapse from low to high (`L'$\rightarrow$`H') state with a constant probability, $p_{Hebb\_pot}$, as illustrated in Fig. \ref{fig:HB_STDP}(a) and described by
\begin{equation}
P_{L \rightarrow H} =
\begin{cases}
p_{Hebb\_pot}, & \text{if $0 < \Delta t \leq t_{Hebb\_pot}$}\\
0, & \text{for all other $\Delta t$}
\end{cases}
\end{equation}
where $P_{L \rightarrow H}$ is the probability of synaptic potentiation. Probabilistic synaptic potentiation is carried out for small time difference between causally related pre- and post-spikes following the Hebbian learning principle that can be summarized as `\textit{neurons that fire together, must wire together}' \citep{lowel1992selection}. Hence, the corresponding timing window is designated as the \textit{Hebbian potentiation} window. On the other hand, probabilistic synaptic depression is carried out for large positive or small negative time difference between the pre- and post-spikes as specified by
\begin{equation}
P_{H \rightarrow L} =
\begin{cases}
p_{antiHebb\_dep}, & \text{if $\Delta t > 0 \  \cap \  \Delta t \geq t_{antiHebb\_dep}$}\\
p_{Hebb\_dep}, & \text{if $t_{Hebb\_dep} \leq \Delta t \leq 0$}\\
0, & \text{for all other $\Delta t$}
\end{cases}
\end{equation}
where $P_{H \rightarrow L}$ is the probability of synaptic depression. We depress the synapse from high to low state with a constant probability, $p_{antiHebb\_dep}$, if the time difference between causally related pre- and post-spikes is larger than $t_{antiHebb\_dep}$, which is anti-Hebbian in nature. Hence, the corresponding STDP timing window is referred to as the \textit{anti-Hebbian depression} window. Anti-Hebbian depression enables the synapses to unlearn features lying outside the neuronal receptive field like noisy background in image patterns. Synaptic depression, in addition, is carried out with a probability, $p_{Hebb\_dep}$, if a pre-spike follows a post-spike and the difference between the respective spike times lies within the negative \textit{Hebbian depression} ([$t_{Hebb\_dep}$, $0$]) window. It is important to note that eHB-STDP contains a dead zone in the STDP timing window, where the switching probability is zero, between the Hebbian potentiation and anti-Hebbian depression windows as depicted in Fig. \ref{fig:HB_STDP}(a). We find that expanding the anti-Hebbian depression window towards the Hebbian potentiation window leads to depression of moderately correlated features in addition to the weakly correlated ones. On the other hand, expanding the Hebbian potentiation window causes the synapses connecting a post-neuron to encode multiple overlapping input features, which negatively impacts the selectivity of the post-neuron and degrades the inference capability of the SNN. The dead zone, in effect, ensures that binary synapses learn and retain strongly correlated input features and unlearn only the weakly correlated ones by facilitating optimal balance between the potentiation and depression updates. We visually demonstrate the significance of dead zone for efficient feature learning using binary fully-connected SNN in \autoref{sec:fully_connected_results}.

Next, we discuss how the eHB-STDP dynamics are adapted for binary synapses connecting inhibitory pre-neurons firing negative spikes. The iHB-STDP dynamics (shown in Fig. \ref{fig:HB_STDP}(b)) are obtained by symmetrically inverting the eHB-STDP dynamics (shown in Fig. \ref{fig:HB_STDP}(a)) about the $\Delta t$ ($t_{post}-t_{pre}$) axis. As a result, the erstwhile potentiation windows are converted to depression windows, and vice versa. According to iHB-STDP, if an inhibitory pre-spike causes the post-neuron to fire and the spike timing difference is smaller than a pre-specified time period, we probabilistically depress the binary synaptic weight. This ensures that the strongly correlated inhibitory (negative) pre-spike modulated by the depressed synaptic weight causes an effective increase in the post-neuronal membrane potential, thereby improving the chances of a post-spike at subsequent time instants. Probabilistic synaptic depression enables a post-neuron to integrate the small positive time difference between an inhibitory pre-spike and the ensuing post-spike, which conforms to the Hebbian learning theory. Probabilistic synaptic potentiation, on the other hand, causes an inhibitory pre-spike modulated by the synaptic weight to lower the post-neuronal membrane potential, thus reducing the chances of a post-spike at subsequent time instants. Hence, it is carried out for large positive time difference (anti-Hebbian in nature) or small negative time difference (Hebbian in nature) between the pre- and post-spikes. The iHB-STDP learning rule for inhibitory pre-neurons effectively incorporates the learning dynamics of eHB-STDP for excitatory pre-neurons by mirroring the potentiation and depression windows about the $\Delta t$ axis.

In this work, we use trace-based technique to estimate spike timing differences as it is commonly adopted for efficient implementation of STDP learning rules \citep{diehl2015unsupervised}. For instance, the positive time difference between a pair of pre- and post-spikes is estimated by generating an exponentially decaying pre-trace (with time constant $\tau_{pre}$) that is reset to unity at the time instant of a pre-spike, and sampling it in the event of a post-spike. Smaller the time difference between the pre- and post-spikes, larger is the sampled pre-trace, and vice versa. Every pre-neuron has a pre-trace that is sampled upon a post-spike to obtain the positive spike timing difference. Likewise, every post-neuron has a post-trace (with time constant $\tau_{post}$) that is sampled upon a pre-spike to obtain the negative spike timing difference. As a result, the eHB-STDP (iHB-STDP) hyperparameters, namely, $t_{Hebb\_pot}$ ($t_{Hebb\_dep}$), $t_{antiHebb\_dep}$ ($t_{antiHebb\_pot}$), and $t_{Hebb\_dep}$ ($t_{Hebb\_pot}$) are mapped to $pre_{Hebb\_pot}$ ($pre_{Hebb\_dep}$), $pre_{antiHebb\_dep}$ ($pre_{antiHebb\_pot}$), and $post_{Hebb\_dep}$ ($post_{Hebb\_pot}$), respectively.

\subsection{Unsupervised Training Methodology for the Convolutional Layers} \label{sec:ReStoCNet_training_methodology}
\begin{figure}[!t]
\centering
\includegraphics[width=6.7in]{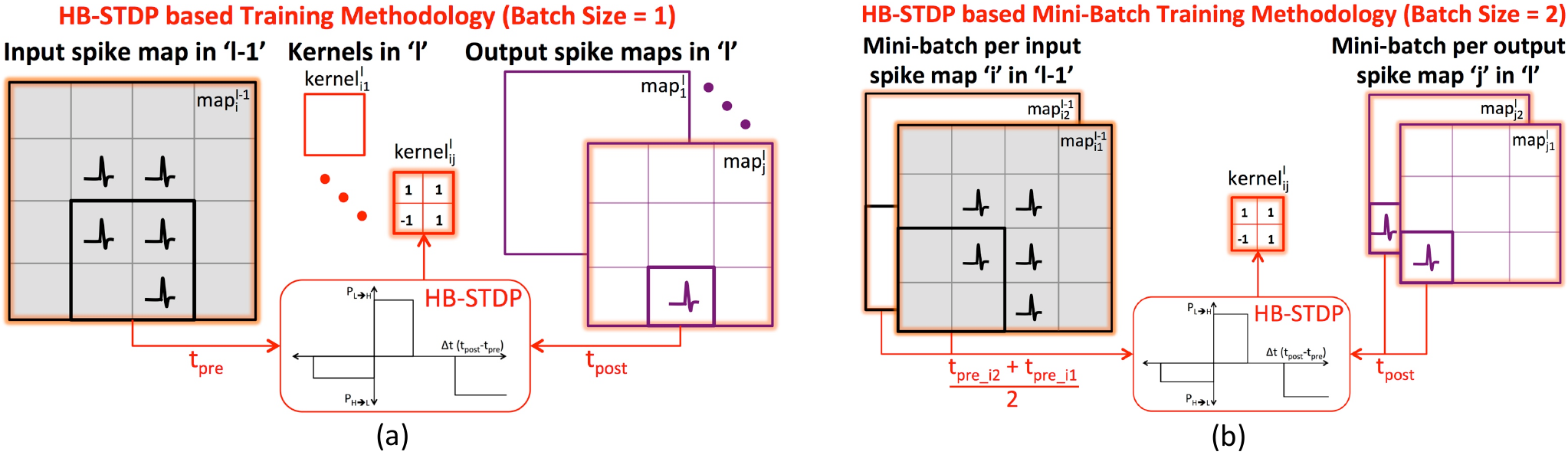}
\caption{(a) Illustration of HB-STDP based unsupervised training methodology for $kernel_{ij}^l$ connecting the $i^{th}$ input spike map in layer `$l-1$' ($map_i^{l-1}$) to the $j^{th}$ output spike map in layer `$l$' ($map_j^l$). The $kernel_{ij}^l$ is updated using HB-STDP based on the spike timing difference between the spiking post-neuron in output $map_j^l$ and the corresponding pre-neurons in input $map_i^{l-1}$. The HB-STDP based weight updates are carried out on all the kernels in layer `$l$' based on the respective input and output maps. (b) Illustration of HB-STDP based mini-batch training methodology for mini-batch size of 2. The $kernel_{ij}^l$ is now updated based on the average spike timing difference between the spiking post-neurons in the output mini-batch ($map_{j1}^l$ and $map_{j2}^l$) and the respective pre-neurons in the input mini-batch ($map_{i1}^{l-1}$ and $map_{i2}^{l-1}$).}
\label{fig:ReStoCNet_train}
\end{figure}
We train the binary kernels forming ReStoCNet in a layer-wise unsupervised manner using the proposed probabilistic e/iHB-STDP learning rule. Consider a $k \times k$ binary kernel ($kernel_{ij}^l$) connecting the $i^{th}$ input spike map in layer `$l-1$' ($map_i^{l-1}$) to the $j^{th}$ output spike map in layer `$l$' ($map_j^l$) as shown in Fig. \ref{fig:ReStoCNet_train}(a). Let us suppose that a post-neuron in the output $map_j^l$ spikes at a particular time instant: the kernel weights are then probabilistically updated based on the time difference between the post-spike and the corresponding $k \times k$ pre-spikes in the input $map_i^{l-1}$. We use the eHB-STDP learning rule for excitatory pre-neurons and iHB-STDP learning rule for inhibitory pre-neurons as described in \autoref{sec:HB-STDP}. If multiple post-neurons in the output $map_j^l$ spike, we update $kernel_{ij}^l$ based on the average spike timing difference between the spiking post-neurons and the respective pre-neurons, which leads to generalized feature learning. However, in order to achieve optimal generalization performance, we average the spike timing differences computed with fixed stride, known as $STDP_{stride}$, over the output $map_j^l$. As an example, for $STDP_{stride}$ of 2, we average the spike timing differences computed between every alternate spiking post-neuron in output $map_j^l$ and the respective pre-neurons. Larger the $STDP_{stride}$, fewer is the number of post-neurons whose spike timing difference estimates are averaged to update the kernel. Consequently, there is loss of generality and added specificity in the features learnt by the kernel for larger $STDP_{stride}$. We experimentally determine the $STDP_{stride}$ for optimal generalization performance that yields the highest test accuracy for a given pattern recognition task.

STDP-based learning is typically performed in an online manner by feeding the input patterns sequentially. STDP-based online learning has been shown to work well particularly for two-layer fully-connected SNNs, where each output or excitatory neuron learns to spike exclusively for a unique class of input patterns by encoding a general input representation in the input to excitatory synaptic weights \citep{diehl2015unsupervised}. Convolutional SNNs, on the other hand, require each kernel to extract features shared across different input classes. In order to enable the kernel to extract general features characterizing different input classes, we perform mini-batch learning following recent works by \citet{lee2018deep} and \citet{ferre2018unsupervised}. The proposed HB-STDP based mini-batch training methodology is illustrated in Fig. \ref{fig:ReStoCNet_train}(b), where the $kernel_{ij}^l$ is now shared by a mini-batch of $i^{th}$ input map in layer `$l-1$' (input mini-batch) and $j^{th}$ output map in layer `$l$' (output mini-batch). We first average the spike timing differences between the spiking post-neurons and the respective pre-neurons, estimated using fixed $STDP_{stride}$, over each output map in the mini-batch to obtain the resultant spike timing difference per output map in the mini-batch. We subsequently average the resultant spike timing differences of the output maps across the mini-batch and probabilistically update $kernel_{ij}^l$ using HB-STDP as shown in Fig. \ref{fig:ReStoCNet_train}(b) for a specific post-neuron in the output mini-batch. At every time instant, the HB-STDP driven mini-batch weight updates are carried out on all the kernels in a given layer. This process is repeated over the entire time duration, $T_{STDP}$, for which the training patterns are presented.

Finally, in order to ensure that different kernels in a layer learn diverse input features, we incorporate the uniform firing threshold adaptation scheme proposed by \citet{lee2018deep} and dropout \citep{srivastava2014dropout} for the output maps. In the beginning of training, the firing threshold of all the post-neurons in every output mini-batch is reset to zero. When a mini-batch of training patterns is presented, multiple post-neurons in an output mini-batch spike and encode definite input features in the kernel weights. We then increase the firing threshold of all the post-neurons in the output mini-batch by an amount $\Delta thresh$, which is specified by
\begin{equation}
\Delta thresh = \beta_{thresh} \times \frac{output\ spike\ count}{output\ map\ size} 
\end{equation}
where $\beta_{thresh}$ is the rate of threshold increase, $output\ spike\ count$ is the number of spikes per output map summed over the mini-batch, and $output\ map\ size$ is the product of the height and width of the output maps. The amount of threshold increase depends on the $output\ spike\ count$ normalized by the $output\ map\ size$ to account for the drop in spiking activity of the output maps across successive convolutional layers due to gradual reduction in the respective sizes. Higher the normalized spiking activity of the output mini-batch, greater is the corresponding increase in its firing threshold, and vice versa. Firing threshold adaptation effectively regulates the spiking activity of the output mini-batch and provides an opportunity for the hitherto dormant output mini-batches to spike and learn, thereby ensuring that no single output mini-batch completely dominates the learning process during a mini-batch training iteration. In addition, we introduce dropout \citep{srivastava2014dropout} for the output maps to achieve diversity in feature learning across successive mini-batch training iterations. At the beginning of every training iteration, we randomly drop a fraction of output mini-batches based on the dropout probability, $p_{drop}$, by forcing the respective spike outputs to zero. Dropout ensures that the same output mini-batch does not spike repeatedly for every training iteration, thereby promoting diversity in feature learning among the kernels in a layer. Once a layer is trained, we propagate the spikes from the input through the trained layers, and update the kernels and firing thresholds of the output maps in the following layer using the presented training methodology. The training process is repeated for all the convolutional layers in ReStoCNet.

\subsection{Supervised Training Methodology for the Fully-connected Layer}
After all the convolutional layers are trained, we pool the respective spike maps using average pooling as detailed in \autoref{sec:ReStoCNet_arch}. We then low-pass filter the spike trains of the pooled maps, by integrating the spike outputs at every time instant and decaying the resultant sum between successive time instants, to obtain their spiking activations as described in \citet{lee2016training, lee2018training} and specified by
\begin{equation}
\begin{array}{l}
pool^l_{lpf}(t) = e^{-\frac{\Delta t_{sim}}{\tau_{lpf}}} \times pool^l_{lpf}(t-\Delta t_{sim}) + pool^l(t) \\
pool^l_{out} = \frac{pool^l_{lpf}(T_{sim})}{T_{sim}}
\end{array} 
\end{equation}
where $pool^l_{lpf}(t)$ is the low-pass filtered output of the pooled spike map $pool^l(t)$ in layer `$l$' at any given time $t$, $\tau_{lpf}$ is the low-pass filter time constant, $\Delta t_{sim}$ is the simulation time-step, $T_{sim}$ is the simulation period for which the input patterns are presented, and $pool^l_{out}$ is the spiking activation of the pooled map in layer `$l$' over the simulation period. The spiking activation thus obtained accounts for the highly non-linear leaky-integrate-and-fire and membrane potential reset dynamics of the spiking neurons in the convolutional layers. The spiking activations of the pooled maps of all the convolutional layers are concatenated and fed to the fully-connected layer, trained using error backpropagation \citep{rumelhart1986learning}, for inference. We use full-precision synaptic weights in the fully-connected layer to comprehensively validate the efficacy of the proposed probabilistic HB-STDP learning rule for training the binary kernels in the convolutional layers. The full-precision synaptic weights can be binarized using algorithms proposed for training binary deep learning networks \citep{courbariaux2015binaryconnect, rastegari2016xnor, hubara2017quantized}. It is important to note that the presented HB-STDP based learning methodology effectuates plasticity by probabilistically switching the binary weights, thereby precluding the need to store the full-precision weights during training. Binarization algorithms for deep learning networks, on the other hand, update the full-precision weights during training, which are subsequently binarized for forward propagation and computing the error gradients.

\section{Results}
We first validate the efficacy of HB-STDP, by visually demonstrating the significance of having distinct potentiation and depression windows separated by a dead zone for efficient feature learning, using two-layer binary fully-connected SNN trained on the MNIST dataset. We then comprehensively evaluate ReStoCNet and the presented HB-STDP based unsupervised mini-batch training methodology on the MNIST and CIFAR-10 datasets. We show that the residual connections are critical to achieving efficient unsupervised learning in deeper convolutional layers and minimizing the accuracy degradation incurred by STDP-trained deep SNNs without residual connections. We use the classification accuracy on the test set and the synaptic memory compression obtained by using binary kernels as the evaluation metrics for ReStoCNet compared to full-precision (32-bit) SNN under iso-accuracy conditions.

\subsection{Two-layer Binary Fully-connected SNN for MNIST Digit Recognition} \label{sec:fully_connected_results}
\begin{figure}[!t]
\centering
\includegraphics[width=6.6in]{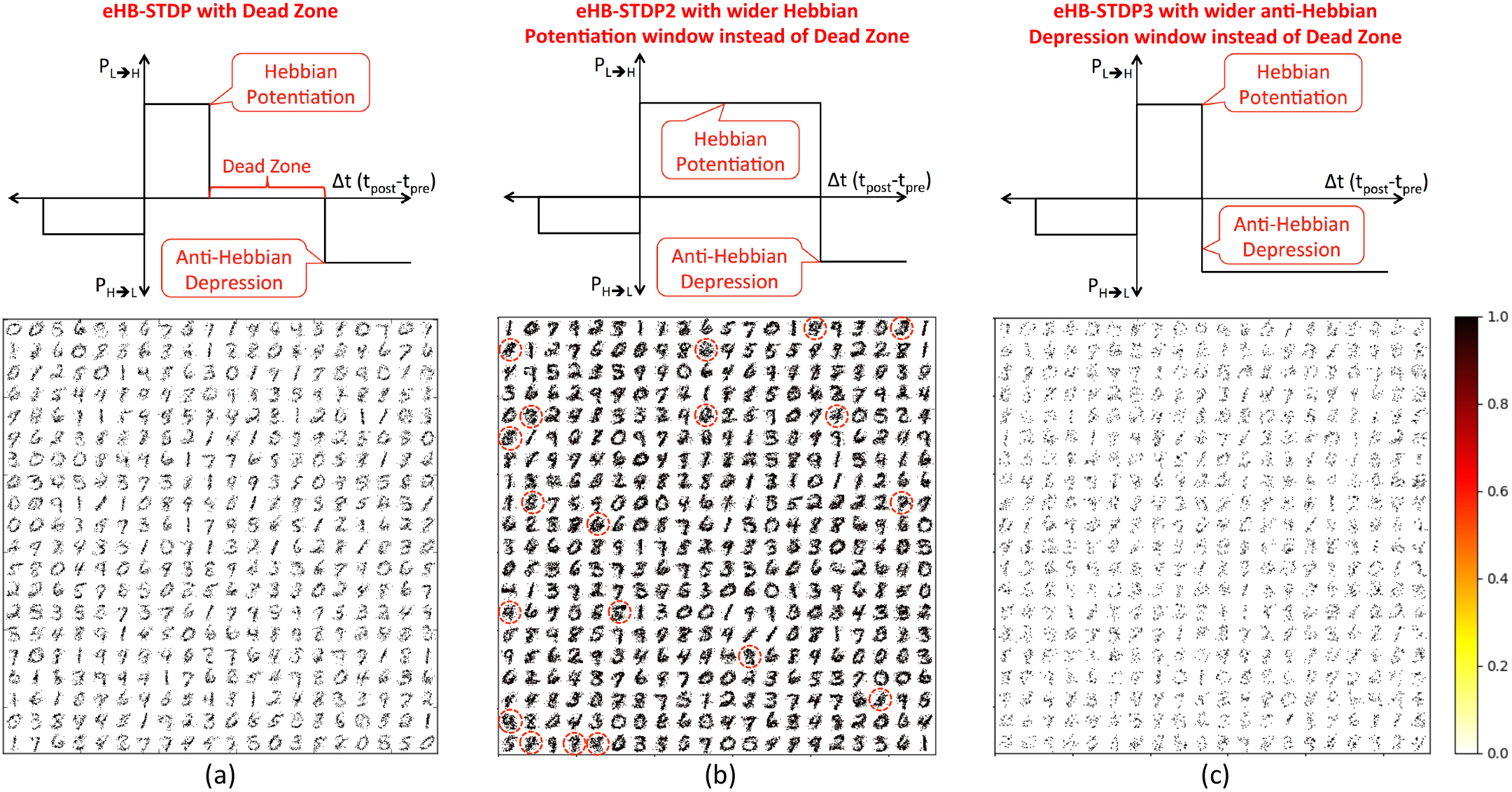}
\caption{MNIST digit representations (re-arranged in 28$\times$28 format) learnt by the synapses connecting the input to each excitatory neuron in a binary fully-connected SNN of 400 neurons (arranged in 20$\times$20 grid). The binary fully-connected SNN is trained using (a) the proposed eHB-STDP containing distinct Hebbian potentiation and anti-Hebbian depression windows separated by a dead zone, (b) eHB-STDP2 where the dead zone is replaced with a wider Hebbian potentiation window, and (c) eHB-STDP3 where the dead zone is replaced with a wider anti-Hebbian depression window.}
\label{fig:fcn_features}
\end{figure}

The binary fully-connected SNN \citep{diehl2015unsupervised} consists of an input layer fully-connected via binary synapses to neurons in the excitatory layer, which are connected in a one-to-one manner to neurons in the subsequent inhibitory layer. Each inhibitory neuron laterally inhibits all the excitatory neurons except the one from which it receives a forward connection. Lateral inhibition facilitates competitive learning and enables each excitatory neuron to spike exclusively and recognize a unique class of input patterns. The input to excitatory synaptic weights are trained using three different configurations of the eHB-STDP learning rule that are enumerated below:
\begin{enumerate}
\item eHB-STDP -- This is the proposed eHB-STDP learning rule containing distinct Hebbian potentiation and anti-Hebbian depression windows separated by a dead zone as shown in Fig. \ref{fig:fcn_features}(a).
\item eHB-STDP2 -- This is a variant of the eHB-STDP learning rule where the dead zone is replaced with a wider Hebbian potentiation window as depicted in Fig. \ref{fig:fcn_features}(b).
\item eHB-STDP3 -- This is an alternative variant of the eHB-STDP rule where the dead zone is replaced with a wider anti-Hebbian depression window as illustrated in Fig. \ref{fig:fcn_features}(c). 
\end{enumerate}
Note that the excitatory$\leftrightarrow$inhibitory synaptic weights are fixed \textit{a priori} and are not subjected to STDP-based learning. We simulated the fully-connected SNN using BRIAN \citep{goodman2008brian}, which is an open-source SNN simulation framework, on the MNIST dataset. The input image pixels are converted to Poisson spike trains firing at a rate constrained between 0 and 63.75$Hz$ depending on the respective pixel intensities for a simulation period of 350$ms$. Note that the simulation time-step is 0.5$ms$. We use the spiking neuronal model detailed in \citet{diehl2015unsupervised} whose parameters are adopted from \citet{jug2012on}. The eHB-STDP hyperparameters used in our simulations are listed in Table \ref{table:fcn_param}.

\begin{figure}[!t]
\centering
\includegraphics[width=5.2in]{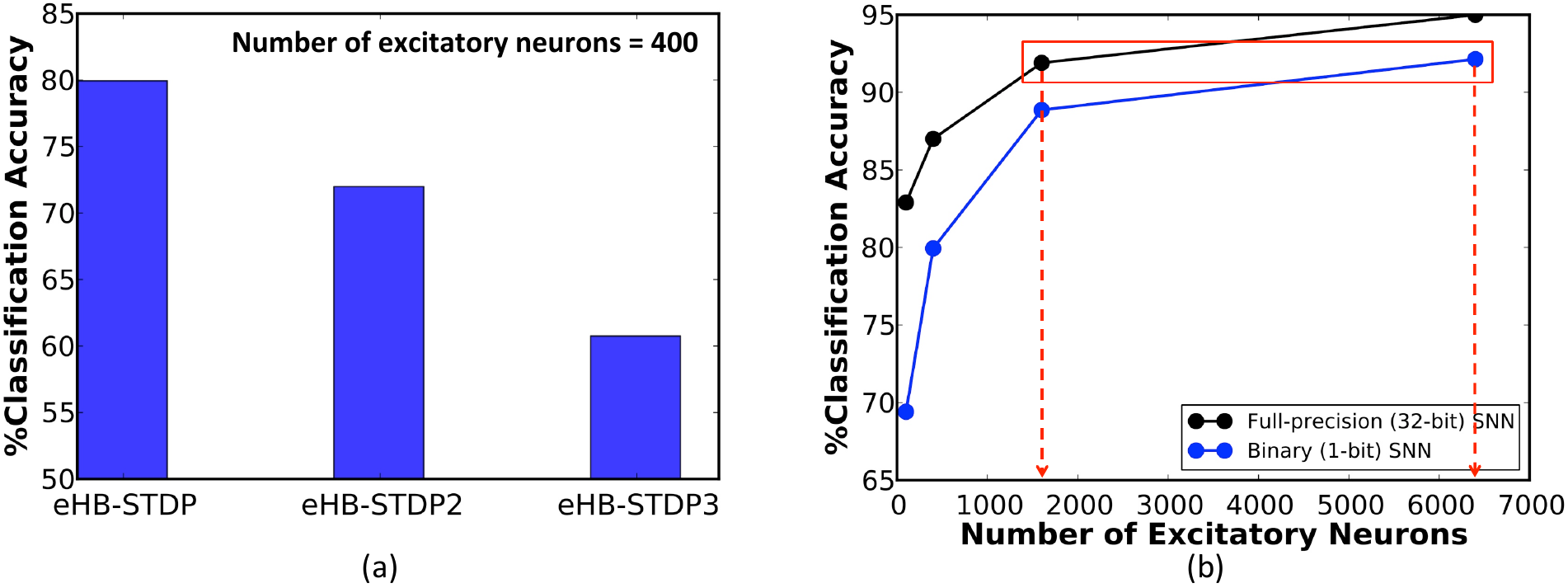}
\caption{(a) Classification accuracy of binary fully-connected SNN of 400 excitatory neurons trained using the three different eHB-STDP configurations illustrated in Fig. \ref{fig:fcn_features}. (b) Classification accuracy of binary fully-connected SNN, trained using the proposed eHB-STDP learning rule, compared to full-precision (32-bit) SNN \citep{diehl2015unsupervised} for different network sizes.}
\label{fig:fcn_accuracy}
\end{figure}
We first train a binary fully-connected SNN of 400 excitatory neurons using the three different eHB-STDP configurations on 3500 MNIST digit patterns. Fig. \ref{fig:fcn_features}(a) illustrates that eHB-STDP causes each excitatory neuron to self-learn general representation of a unique digit in the input to excitatory synaptic weights. On the other hand, eHB-STDP2, with a wider Hebbian potentiation window instead of the dead zone, causes certain excitatory neurons to self-learn overlapping input representations as highlighted in Fig. \ref{fig:fcn_features}(b). Overlapping input representations negatively impact the selective spiking behavior of the excitatory neurons for specific input classes and degrade the recognition capability of the SNN. The final eHB-STDP configuration, eHB-STDP3, leads to insufficient representation learning as depicted in Fig. \ref{fig:fcn_features}(c) due to the dominance of synaptic depression over synaptic potentiation weight updates. Thus, the proposed eHB-STDP learning rule offers superior representation learning capability compared to the explored variants by maintaining optimal balance between the potentiation and depression weight updates. This is further corroborated by the accuracy results shown in Fig. \ref{fig:fcn_accuracy}(a), which is evaluated as explained below. At the end of eHB-STDP based training, each excitatory neuron is tagged as having learnt the class of input patterns for which it spiked the most during the training phase. A test pattern is predicted to belong to the class (or tag) represented by the group of neurons with the highest average spike count over the simulation period. The binary fully-connected SNN of 400 neurons trained using eHB-STDP yielded 79.94\% accuracy on the MNIST test set, which is higher by $>$8\% compared to that achieved using the remaining eHB-STDP variants. The accuracy can be further improved by increasing the number of excitatory neurons as shown in Fig. \ref{fig:fcn_accuracy}(b). We now estimate the $synaptic\ memory\ compression$ offered by the binary SNN compared to full-precision (32-bit) SNN, which is specified by
\begin{equation}
synaptic\ memory\ compression = \frac{\#input\ neurons \times \#excitatory\ neurons_{full-precisionSNN} \times 32}{\#input\ neurons \times \#excitatory\ neurons_{binarySNN} \times 1}
\end{equation}
where $\#input\ neurons$ is 784 for the MNIST dataset. Fig. \ref{fig:fcn_accuracy}(b) indicates that binary SNN of 6400 neurons offers comparable accuracy ($\sim$92\%) to that provided by full-precision (32-bit) SNN of 1600 neurons \citep{diehl2015unsupervised}, leading to 8$\times$ synaptic memory compression under iso-accuracy conditions. Note that the accuracy of $\sim$92\% is higher than that reported in related works for binary fully-connected SNN, trained using probabilistic STDP-based learning rules, as shown in Table \ref{table:binary_fcn_mnist}. However, the fully-connected SNN introduces scalability issues as the network depth is increased due to explosion in the number of trainable parameters. We demonstrate ReStoCNet, which is a scalable multilayer convolutional SNN composed of binary kernels, trained using the optimal e/iHB-STDP based unsupervised mini-batch training methodology.

\subsection{ReStoCNet for MNIST Digit Recognition} \label{sec:ReStoCNet_MNIST}
The MNIST dataset contains 60,000 training patterns and 10,000 test patterns of handwritten digits that are stored as 28$\times$28 Greyscale images. In this work, we developed a custom simulation framework using Pytorch \citep{paszke2017automatic} to evaluate ReStoCNet and the presented HB-STDP based unsupervised training methodology. The simulation parameters for the Leaky-Integrate-and-Fire (LIF) neuron in the convolutional layers and the Integrate-and-Fire (IF) neuron in the spatial pooling layers are shown in Table \ref{table:ReStoCNet_conv}. The binary kernels in every convolutional layer are initialized to logic high state ($w_{high}$) with a probability, $p_{high}$, which is specified by
\begin{equation}
p_{high} = \sqrt{\frac{\alpha_{weight\_init}}{fan\_in + fan\_out}}
\end{equation}  
where $\alpha_{weight\_init}$ is the proportionality constant controlling $p_{high}$, and $fan\_in$ and $fan\_out$ are the total number of input and output synaptic weights, respectively, for a given convolutional layer. The remaining kernel weights in the convolutional layer are initialized to logic low state ($w_{low}$). The firing threshold of the LIF neurons in every convolutional layer are initialized to zero.

\begin{figure}[!t]
\centering
\includegraphics[width=2.8in]{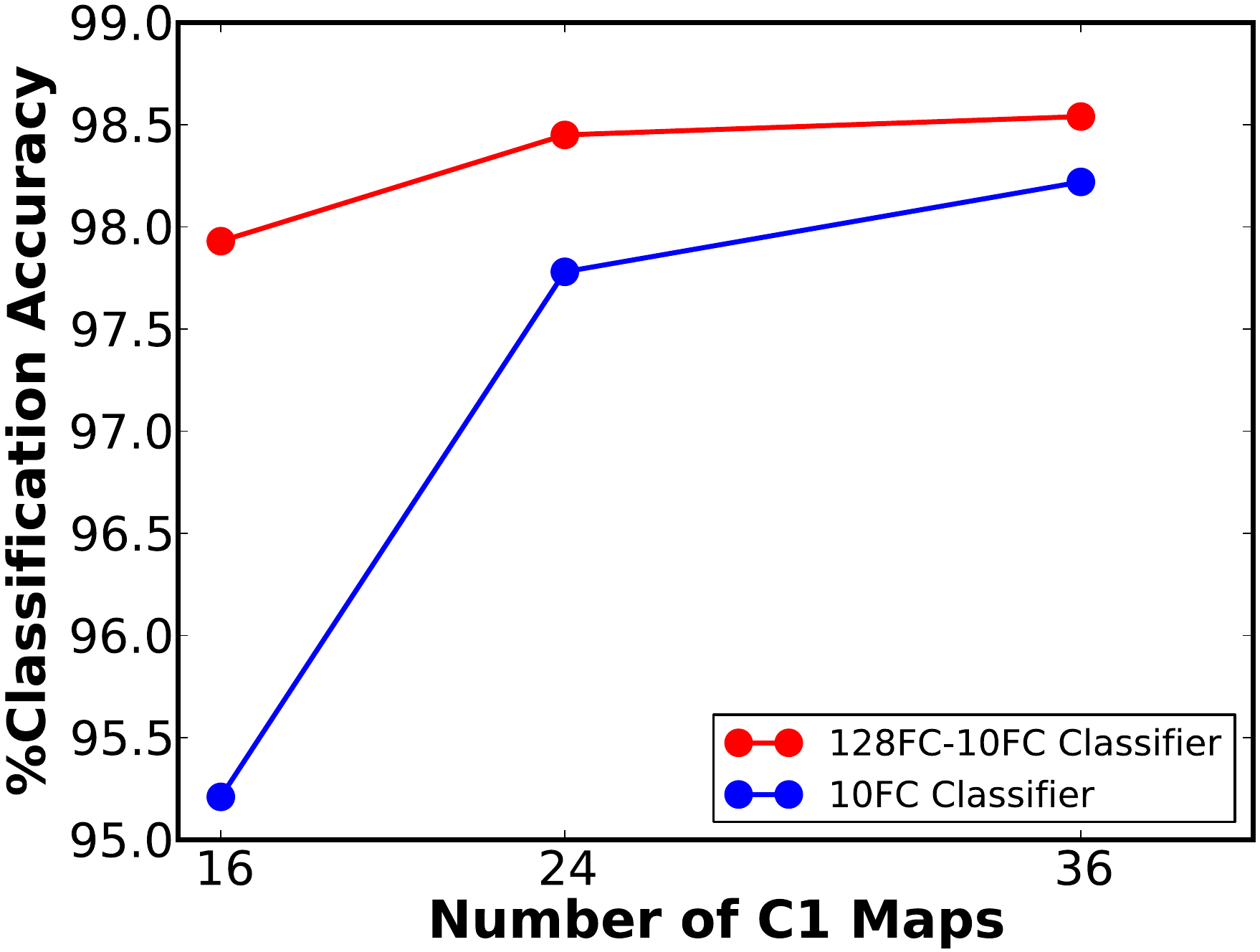}
\caption{Classification accuracy of ReStoCNet, composed of single convolutional layer followed by a pooling layer and one or more fully-connected layers, versus the number of output (C1) maps, on the MNIST test set.}
\label{fig:c1_results_mnist}
\end{figure}

We first simulated a 16C3-2P-10FC ReStoCNet, composed of single convolutional layer with 16 maps and 3$\times$3 binary kernels followed by pooling layer whose spiking activations are directly fed to the final softmax layer. The input image pixels are mapped to excitatory pre-neurons firing at a rate constrained between 0 and 200$Hz$ depending on the corresponding pixel intensities. The eHB-STDP model parameters are provided in Table \ref{table:ReStoCNet_conv}. We trained the convolutional layer in ReStoCNet using 2000 MNIST digit patterns with a mini-batch size of 200. We thereafter fed the entire training dataset to ReStoCNet, spatially pooled the spike maps of the convolutional layer, and low-pass filtered the pooled spike trains over a simulation period of 100$ms$ to estimate their spiking activations. The pooling layer spiking activations are passed on to the fully-connected softmax layer, which is trained using the Adam optimizer \citep{kingma2014adam} and cross-entropy loss function for 100 epochs. The training parameters used for the fully-connected layer are mentioned in Table \ref{table:ReStoCNet_fcn}. The shallow ReStoCNet yielded an accuracy of 95.21\% on the MNIST test set, which increased to 98.22\% for a wider 36C3-2P-10FC ReStoCNet in which the convolutional layer is trained using 10,000 MNIST digit patterns. Further improvement in accuracy is obtained by augmenting the classifier in ReStoCNet with an additional fully-connected layer of 128 neurons prior to the softmax output layer as shown in Fig. \ref{fig:c1_results_mnist}, which indicates that 36C3-2P-128FC-10FC ReStoCNet offers an improved accuracy of 98.54\% on the MNIST test set. Note that we did not simulate deep ReStoCNets for MNIST digit recognition since the shallow networks yield $>$98\% accuracy, and that any further increase in the depth of STDP-trained convolutional layers would not provide commensurate improvements in the classification accuracy.

\subsection{ReStoCNet for CIFAR-10 Image Recognition} \label{sec:ReStoCNet_CIFAR}
\begin{figure}[!t]
\centering
\includegraphics[width=4.0in]{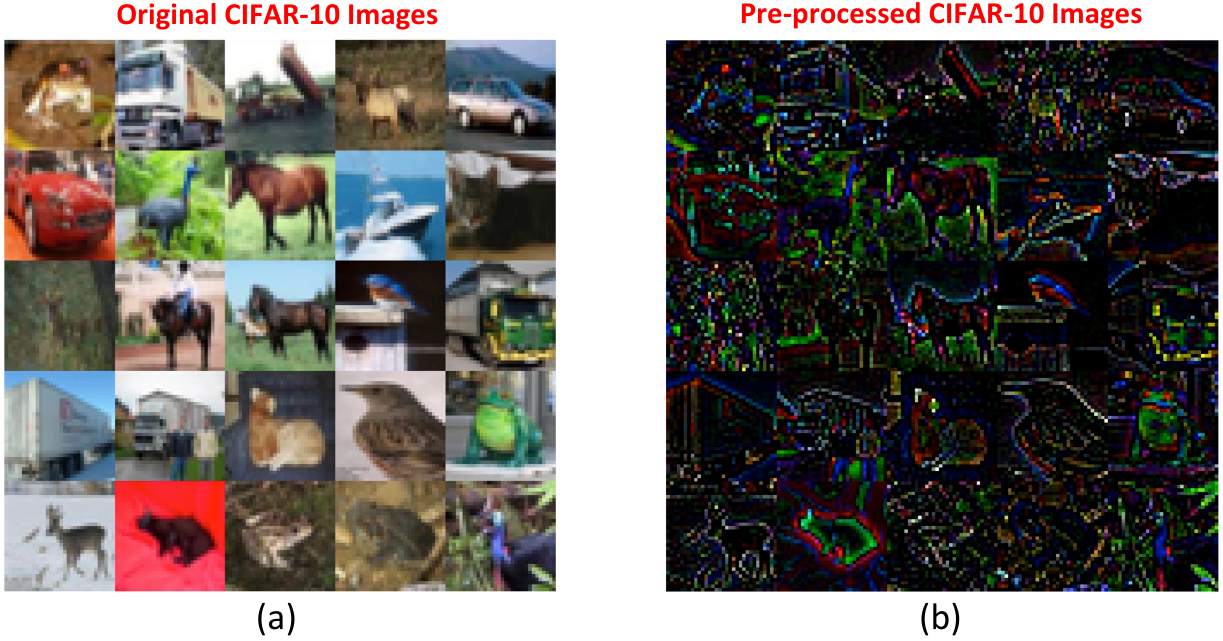}
\caption{(a) Original 32$\times$32$\times$3 CIFAR-10 images. (b) CIFAR-10 images pre-processed using global contrast normalization followed by ZCA whitening \citep{krizhevsky2009learning}.}
\label{fig:cifar10_montage}
\end{figure}

The CIFAR-10 dataset contains 50,000 training images and 10,000 test images, 32$\times$32$\times$3 in dimension, spanning 10 output classes. We pre-processed the CIFAR-10 images using global contrast normalization followed by ZCA whitening \citep{krizhevsky2009learning}. Global contrast normalization is performed by subtracting and scaling the pixel intensities of each input channel by the corresponding mean and standard deviation computed over the training set. The normalized image is then transformed by multiplying with whitening filters as explained in \citet{krizhevsky2009learning}, which enables a network to learn higher-order pixel correlations. Fig. \ref{fig:cifar10_montage} illustrates a few original and pre-processed images from the CIFAR-10 dataset. The simulation parameters used for training the convolutional layers are provided in Table \ref{table:ReStoCNet_conv} while those used for training the fully-connected layer are listed in Table \ref{table:ReStoCNet_fcn}. The binary kernels and firing thresholds of the convolutional layers are initialized as described in \autoref{sec:ReStoCNet_MNIST}.

\begin{figure}[!t]
\centering
\includegraphics[width=6.9in]{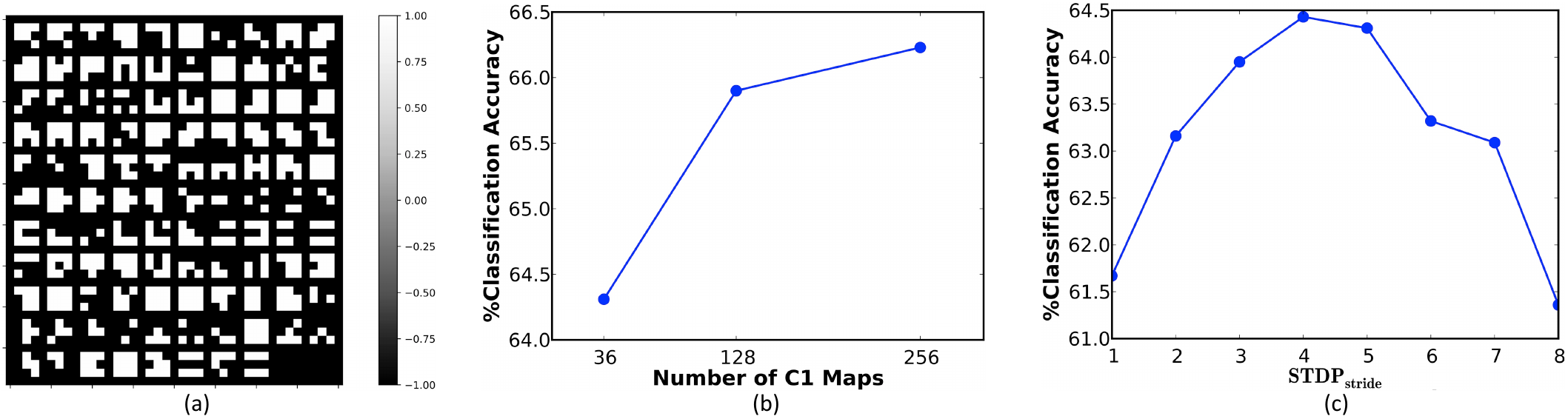}
\caption{(a) Binary kernels (3$\times$3 in size) of ReStoCNet-1 (36C3-2P-1024FC-10FC ReStoCNet), trained using e/iHB-STDP based unsupervised training methodology, on 5000 images from the CIFAR-10 dataset. (b) Classification accuracy of ReStoCNet versus the number of convolutional maps. (c) Classification accuracy of ReStoCNet-1 versus the $STDP_{stride}$ used to compute the average spike timing difference of the spiking post-neurons in the convolutional maps.}
\label{fig:c1_results}
\end{figure}
In our first experiment, we simulated a 36C3-2P-1024FC-10FC ReStoCNet, designated as ReStoCNet-1, consisting of a single convolutional layer with 36 maps and 3$\times$3 binary kernels followed by fully-connected layer containing 1024 ReLU neurons and a final softmax layer with 10 output neurons. The pre-processed CIFAR-10 images are composed of pixels with positive and negative intensities, which are respectively mapped to excitatory and inhibitory pre-neurons firing at a rate constrained between 0 and 200$Hz$ depending on the absolute value of the corresponding pixel intensities. The e/iHB-STDP model parameters are listed in Table \ref{table:ReStoCNet_conv}. Note that the e/iHB-STDP switching probability is set to zero in the negative STDP timing window to facilitate optimal balance between the potentiation and depression updates for a smaller 3$\times$3 kernel shared by 32$\times$32 pre-neurons in the input map and 30$\times$30 post-neurons in the convolutional map. The binary kernels in ReStoCNet-1 are trained using 5000 images, with mini-batch size of 200, for simulation period of 25$ms$ per mini-batch training iteration. Note that we used a simulation time-step of 1$ms$. Fig. \ref{fig:c1_results}(a) illustrates the low-level input features self-learnt by the binary kernels, enabled by the e/iHB-STDP based unsupervised training methodology. The shallow ReStoCNet-1, wherein the fully-connected layer is trained on the entire dataset, yielded 64.31\% test accuracy that is higher than an accuracy of 59.42\% obtained using randomly initialized binary kernels and zero firing thresholds in the convolutional layer. In order to determine if accuracy loss is incurred as a result of using binary kernels, we trained ReStoCNet-1 composed of full-precision (32-bit) kernels using standard exponential STDP rule \citep{song2000competitive} with learning rate of 0.01. ReStoCNet-1 with full-precision kernels provided 64.30\% test accuracy, which is comparable to that obtained using binary kernels. Fig. \ref{fig:c1_results}(b) shows that the test accuracy improves with the number of maps in the convolutional layer. As explained in \autoref{sec:ReStoCNet_training_methodology}, the classification accuracy of ReStoCNet has a strong dependence on the chosen $STDP_{stride}$ used for computing the average spike timing difference of the spiking post-neurons in the convolutional maps. Fig. \ref{fig:c1_results}(c) indicates that the accuracy of ReStoCNet-1 degrades for $STDP_{stride}$ smaller than 4 or greater than 5. If the $STDP_{stride}$ is small, the binary kernels are updated based on the spike timing difference averaged over large number of spiking post-neurons in the convolutional maps, leading to degradation in the learnt features. On the contrary, if the $STDP_{stride}$ is large, the binary kernels are updated based on the spike timing difference estimates of few post-neurons, leading to loss of generality in the learnt features. We use the optimal $STDP_{stride}$ of 5 for all the ReStoCNet experiments presented in this work.

\begin{figure}[!t]
\centering
\includegraphics[width=5.1in]{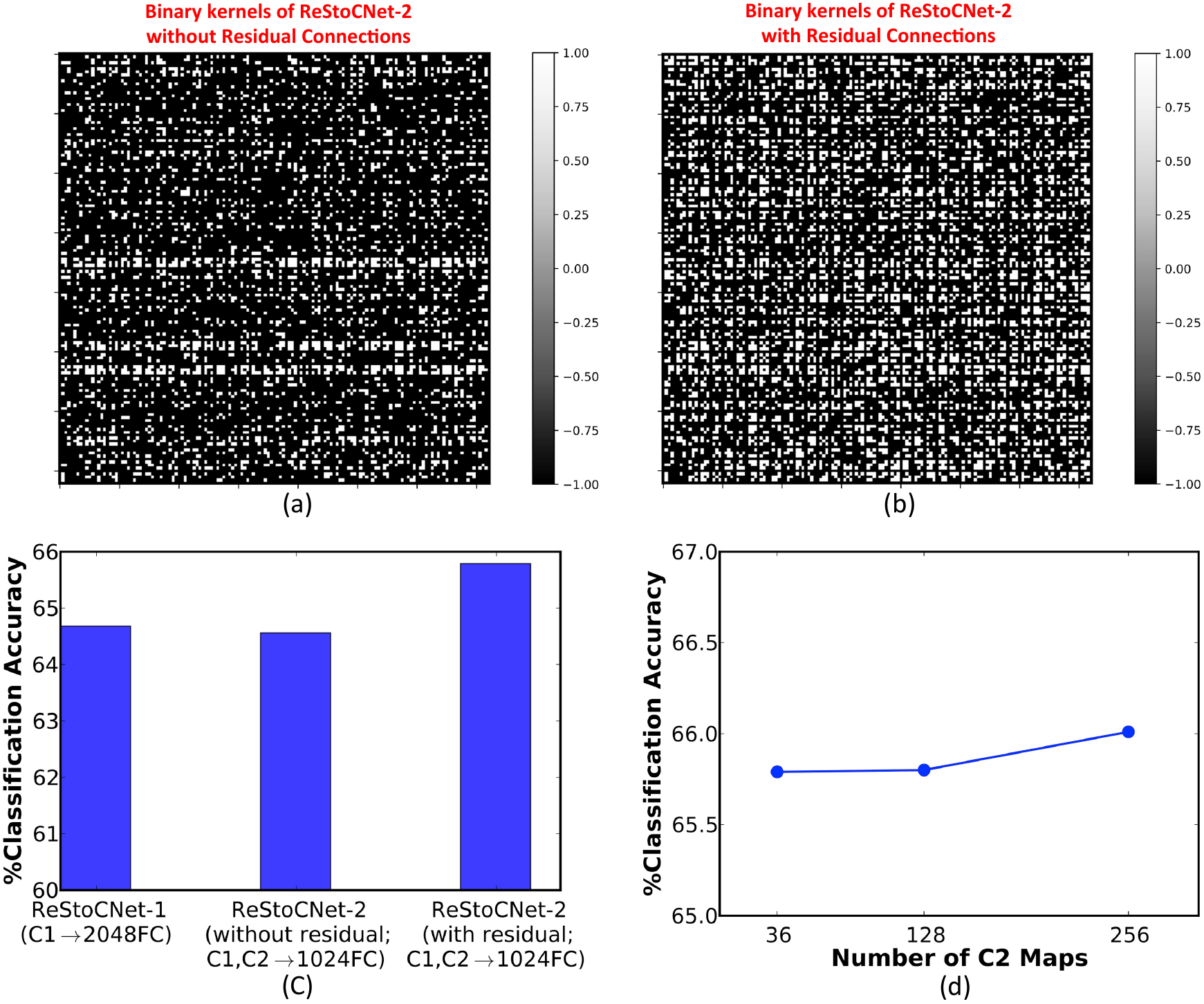}
\caption{Binary kernels (in second convolutional layer) of ReStoCNet-2 (36C3-36C3-2P-1024FC-10FC ReStoCNet) (a) without residual connections, and (b) with residual connections from the input to second convolutional layer. (c) Classification accuracy of ReStoCNet-2 with and without residual connections compared to that provided by ReStoCNet-1. (d) Classification accuracy of ReStoCNet-2 (with residual connections) versus the number of output maps in the second convolutional layer.}
\label{fig:c2_results}
\end{figure}
Next, we simulated a 36C3-36C3-2P-1024FC-10FC ReStoCNet, designated as ReStoCNet-2, composed of two convolutional layers, each with 36 maps and 3$\times$3 binary kernels. The first convolutional layer is trained as described in the previous paragraph. The binary kernels and firing thresholds of the second convolutional layer are trained using a different subset of 5000 CIFAR-10 images with a mini-batch size of 200. Note that the e/iHB-STDP hyperparameters are similar for both the convolutional layers except the synaptic switching probabilities, which are scaled down for the second convolutional layer as shown in Table \ref{table:ReStoCNet_conv}. The lower switching probabilities for the second convolutional layer accounts for the fact that every constituting post-neuron receives weighted input from 36 maps each in the residual and direct paths while a post-neuron in the first convolutional layer receives weighted input from just the 3 maps in the input layer. We simulated two versions of ReStoCNet-2: one without residual connections and the other with residual connections from the input to second convolutional layer. Fig. \ref{fig:c2_results} shows that ReStoCNet-2 with residual connections learns diverse high-level input features compared to the one without residual connections. As a result, ReStoCNet-2 with residual connections yielded 65.79\% accuracy, which is roughly 1.5\% higher than that provided by ReStoCNet-2 without residual connections as well as ReStoCNet-1. This begs the following question: \textit{is ReStoCNet-2 yielding higher accuracy that ReStoCNet-1 just due to increased number of synaptic weights in the fully-connected layer as a consequence of concatenating the pooled spiking activations of both the convolutional layers?} To answer this question, we compare ReStoCNet-2, in which the spiking activations of the 72 pooled maps are fed to a fully-connected layer of 1024 neurons, with ReStoCNet-1 in which the spiking activations of the 36 pooled maps are fed to a larger fully-connected layer of 2048 neurons. Fig. \ref{fig:c2_results}(c) indicates that ReStoCNet-2 offers higher accuracy than that provided by ReStoCNet-1 with 2048 neurons in the fully-connected layer, which is a testament to the improved feature learning capability of the second convolutional layer in the presence of residual inputs. Fig. \ref{fig:c2_results}(d) shows that ReStoCNet-2 provides only modest improvement in accuracy as the number of output maps is increased in the second convolutional layer. The accuracy limitation is caused by the inability of the unsupervised training methodology to effectively optimize an over-parameterized network.

\begin{figure}[!b]
\centering
\includegraphics[width=5.5in]{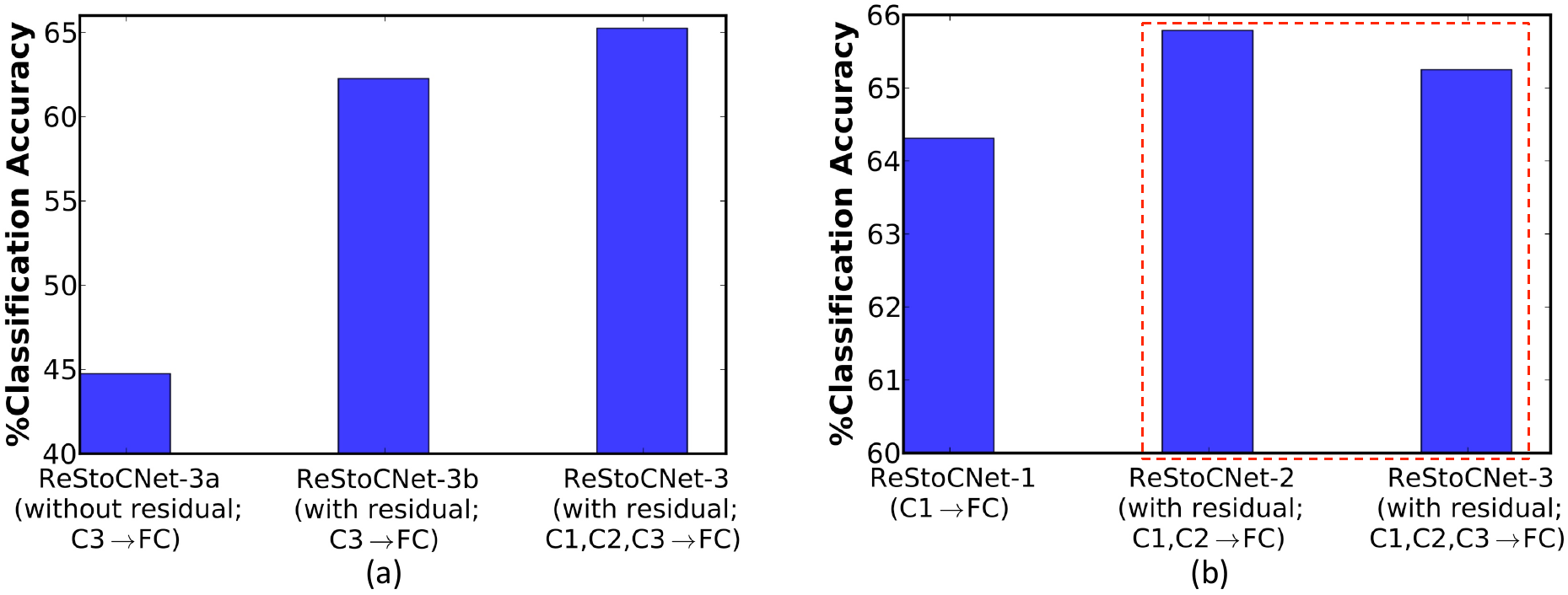}
\caption{(a) Classification accuracy of three different ReStoCNet-3 (36C3-36C3-36C3-2P-1024FC-10FC) configurations on the CIFAR-10 test set. (b) Comparison between the classification accuracy of different ReStoCNet configurations presented in this work.}
\label{fig:c3_results}
\end{figure}
Finally, we evaluated a deeper 36C3-36C3-36C3-2P-1024FC-10FC ReStoCNet, referred to as ReStoCNet-3, composed of three convolutional layers as depicted in Fig. \ref{fig:ReStoCNet_arch}. We inverted the residual inputs to the third convolutional layer to ensure diversity in the residual maps received by the second and third layers from the input layer. We trained the third convolutional layer with the same hyperparameters (shown in Table \ref{table:ReStoCNet_conv}) as those used for training the second convolutional layer, albeit on a different subset of 5000 images from the CIFAR-10 dataset. In addition to ReStoCNet-3 (with residual connections), wherein the pooled spiking activations of all the convolutional layers are used for inference, we simulated the following variants to demonstrate the significance of residual connections for the scalability of deep SNNs:
\begin{enumerate}
\item ReStoCNet-3a -- This is a variant of ReStoCNet-3 without residual inputs to the third convolutional layer. In addition, the pooled spiking activations of only the third convolutional layer are fed to the fully-connected layer for inference.
\item ReStoCNet-3b -- This is a variant of ReStoCNet-3 with residual inputs to the third convolutional layer, wherein the pooled spiking activations of only the third convolutional layer are used for inference.
\end{enumerate}
ReStoCNet-3a, devoid of residual connections, yielded 44.75\% accuracy on the CIFAR-10 test set, which is 17.5\% lower compared to an accuracy of 62.26\% provided by ReStoCNet-3b with residual connections as shown in Fig. \ref{fig:c3_results}(a). The higher accuracy of ReStoCNet-3b can be directly attributed to its improved feature learning capability, rendered possible by the residual inputs feeding into the third convolutional layer. The optimal ReStoCNet-3 configuration (with residual connections), wherein the pooled spiking activations of all the convolutional layers are used for inference, offered 65.25\% accuracy, which is only comparable to an accuracy of 65.79\% provided by ReStoCNet-2 as shown in Fig. \ref{fig:c3_results}(b).

\begin{figure}[!t]
\centering
\includegraphics[width=6.9in]{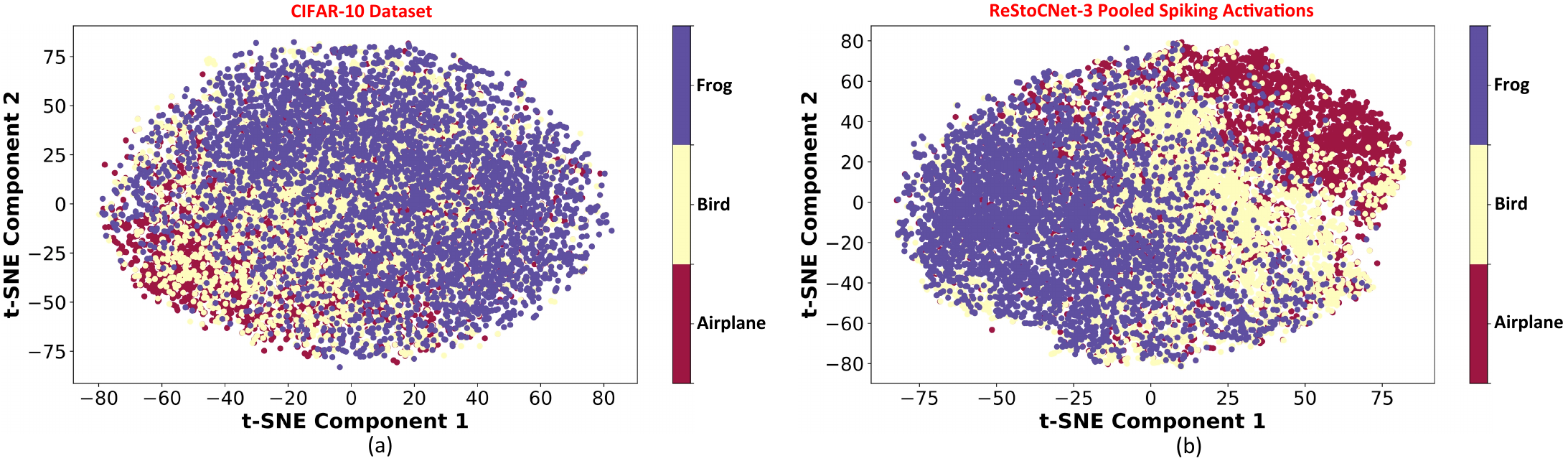}
\caption{(a) Scatter plot of the first two t-SNE components for 15,000 training images from the CIFAR-10 dataset from the classes: airplane, bird, and frog. (b) t-SNE scatter plot of the pooled spiking activations of the convolutional layers in ReStoCNet-3 for the corresponding training images.}
\label{fig:c3_tsne}
\end{figure}

Our analysis on ReStoCNet, trained using the e/iHB-STDP based unsupervised training methodology, offers the following key insights. First, it shows that the residual connections are critical for the scalability of deep SNNs. Second, it reveals that the maximum achievable accuracy is limited by the STDP-based unsupervised training methodology as further corroborated by Fig. \ref{fig:c3_tsne}, which illustrates the unsupervised clustering capability of ReStoCNet-3 for different training images from the CIFAR-10 dataset. In order to visualize the efficiency of unsupervised clustering offered by ReStoCNet-3, we reduce the dimension of the pooled spiking activations of the convolutional layers using Principal Component Analysis (PCA) followed by t-Distributed Stochastic Neighbor Embedding (t-SNE) \citep{maaten2008visualizing}, and plot the first two t-SNE components for the training images. The t-SNE dimensionality reduction technique computes pair-wise similarities between the data points (images) in the high-dimensional space and projects them to a low-dimensional space that preserves the measured similarities. We refer the readers to \citet{maaten2008visualizing} for a review of the t-SNE algorithm for visualizing high-dimensional input data. Fig. \ref{fig:c3_tsne}(a) shows the t-SNE scatter plot for 15,000 training images spanning three different classes from the CIFAR-10 dataset, namely, airplane, bird, and frog. The primary objective of any machine learning model is to cluster the images per class together while ensuring sufficient separation among different classes. The t-SNE scatter plot of the pooled spiking activations of ReStoCNet-3 (shown in Fig. \ref{fig:c3_tsne}(b)) indicates that, although distinct clusters are formed for the images in each class, there exists considerable overlap among different image clusters.

\section{Discussion}
\subsection{Comparison with Related Works} \label{sec:ReStoCNet_comparison}
We compare ReStoCNet with convolutional SNNs, which employ unsupervised training methodology for the convolutional layers and supervised training algorithms like error backpropagation for the fully-connected layer, using classification accuracy (on the test set) and kernel memory compression as the evaluation metrics. The memory compression offered by ReStoCNet as a result of using binary kernels in the convolutional layers, referred to as kernel memory compression, is computed as specified by
\begin{equation}
kernel\ memory\ compression = \frac{N_{baseline} \times ksize_{baseline} \times ksize_{baseline} \times nbits_{full\_precision}}{N_{ReStoCNet} \times ksize_{ReStoCNet} \times ksize_{ReStoCNet} \times nbits_{binary}} 
\end{equation}
where $N_{ReStoCNet}$ ($N_{baseline}$) and $ksize_{ReStoCNet}$ ($ksize_{baseline}$) are the number of kernels and kernel size, respectively, in ReStoCNet (baseline convolutional SNN used for comparison), and $nbits_{binary}$ and $nbits_{full\_precision}$ are the hardware bit-precision required for storing the binary and full-precision kernels, which are set to 2-bits and 32-bits, respectively. Note that the binary kernels in ReStoCNet require storage capacity of 2-bits per synaptic weight since they are constrained to binary states $-1$ and $+1$. Table \ref{table:ReStoCNet_mnist} shows that the classification accuracy offered by ReStoCNet for MNIST digit recognition is comparable to that reported for convolutional SNNs composed of full-precision kernels trained using unsupervised learning methodologies. Specifically, a 36C3-2P-128FC-10FC ReStoCNet offers 98.54\% accuracy on the MNIST test set, which compares favorably with that (98.36\%) provided by the convolutional SNN presented in \citet{tavanaei2017multi}, composed of single convolutional layer with 32 maps and 5$\times$5 full-precision kernels trained using STDP. The proposed ReStoCNet offers 39.5$\times$ kernel memory compression by virtue of using smaller 3$\times$3 binary kernels under iso-accuracy conditions for MNIST digit recognition. On the contrary, very few works have benchmarked convolutional SNNs, trained using unsupervised learning algorithms, on the CIFAR-10 dataset. \citet{panda2016unsupervised} proposed spike-based convolutional Auto-Encoders, where the kernels in every convolutional layer are trained in an unsupervised manner using error backpropagation to regenerate the input spike patterns. \citet{ferre2018unsupervised} presented convolutional SNN (without residual connections), where the kernels are trained using a simple Hebbian STDP learning rule. Table \ref{table:ReStoCNet_cifar} shows that ReStoCNet provides 4-5\% lower accuracy than that reported in both the related works. In particular, a 256C3-2P-1024FC-10FC ReStoCNet yields 4.97\% lower accuracy than that provided by the 64C7-8P-512FC-512FC-10FC convolutional SNN \citep{ferre2018unsupervised} while offering 21.7$\times$ kernel memory compression. Note that the convolutional SNN presented in \citet{ferre2018unsupervised} is simulated by single-step forward propagation using input rates while ReStoCNet is simulated using input spike trains over multiple time-steps.

Finally, we note that deep learning Binary Neural Networks (BNNs) \citep{courbariaux2015binaryconnect, rastegari2016xnor, hubara2017quantized}, which use binary activations for the neurons in every layer except the input and output layers and binary weights, have been demonstrated to yield superior classification accuracy than that provided by ReStoCNet. Nevertheless, ReStoCNet offers the following advantages over BNNs. First, ReStoCNet is inherently suited for processing spatiotemporal spike trains from event-based audio and vision sensors as shown by \citet{stromatias2017event} for convolutional SNNs with full-precision weights since it computes with static image pixels mapped to spike trains. BNNs, on the contrary, use real-valued pixel intensities for the input layer. Second, ReStoCNet is amenable for efficient implementation in event-driven asynchronous neuromorphic hardware platforms like IBM \textit{TrueNorth} \citep{merolla2014million} and Intel \textit{Loihi} \citep{davies2018loihi} since it uses \{$0$, $1$\} for the outputs of the spiking neurons in every convolutional layer. The weighted sum of the input spikes with the synaptic weights in the convolutional layers needs to be computed only in the event of a spike fired by the corresponding input neurons. In addition, only the sparse spiking events need to be transmitted between the layers. The event-driven computing capability offered by ReStoCNet can be exploited to achieve higher energy efficiency in neuromorphic hardware implementations by minimizing the computation and communication energy in the absence of spiking events. BNNs, on the other hand, use \{$1$, $-1$\} for the neuronal activations and either \{$1$, $-1$\} \citep{courbariaux2015binaryconnect} or \{$\alpha$, $-\alpha$\} \citep{rastegari2016xnor} where $\alpha$ is a layer-wise scaling factor for the weights to achieve good accuracy and stable training convergence \citep{pfeiffer2018deep}. Hence, the computation of the weighted input sum and communication of the binarized neuronal activations need to be carried out for all the neurons in every layer in a synchronous manner, which is in contrast to the event-based asynchronous computing capability provided by ReStoCNet. Last, ReStoCNet offers a memory-efficient solution for enabling on-chip intelligence in resource-constrained battery-powered Internet of Things (IoT) edge devices since the binary kernels are trained using probabilistic-STDP based local learning rule that can be efficiently implemented on-chip. Learning is achieved by probabilistically switching the binary kernel weights between the allowed states based on spike timing, which precludes the need for storing the full-precision weights and enhances the memory efficiency during training. BNNs, on the other hand, are trained using error backpropagation algorithms that update the full-precision weights based on the backpropagated error gradients and binarize the modified weights for forward propagation and computing the error gradients. Thus, ReStoCNet provides a promising alternative for energy- and memory-efficient computing during both training and inference in IoT edge devices, for instance, surveillance cameras, which produce large volumes of real-time data. It is inefficient for these devices to continuously offload raw/compressed data to the cloud for training. This is because the sheer volume of generated data could exceed the bandwidth available for transmitting them to the cloud. Alternatively, there could be connectivity issues restricting communication between the edge and the cloud. In addition, there are also security and data privacy issues that need to be addressed while sending (receiving) data to (from) the cloud. Hence, it is highly desirable to equip the edge devices with on-chip intelligence so that they can learn from real-time input data and invoke the cloud occasionally to update the on-chip trained weights using more complex algorithms. The proposed approach is also suited for building intelligent autonomous systems like robots and self-flying drones. For example, it is beneficial to embed on-chip learning in autonomous robots used for disaster relief operations that enables them to navigate obstacles and scour the disaster site for survivors. In the instance of self-flying drones used for reconnaissance operations, on-chip intelligence can enable them to effectively navigate the enemy territory and improve the chances of a successful mission.

The classification accuracy of ReStoCNet for complex applications could be improved by augmenting the layer-wise unsupervised training methodology with a global supervised training mechanism. Recent works have proposed error backpropagation algorithms for the supervised training of SNNs \citep{lee2016training, panda2016unsupervised, wu2018spatio, mostafa2018supervised, lee2018training, jin2018hybrid}. However, the backpropagation algorithms for SNNs, some of which backpropagate errors at multiple time-steps, are computationally prohibitive and prone to unstable convergence behaviors \citep{lee2018training}. In this regard, \citet{neftci2017event} proposed event-driven random backpropagation that prevents the need for calculating and backpropagating precise error gradients. Future works could explore a hybrid unsupervised (local) and supervised (global) training methodology for ReStoCNet to obtain favorable trade-offs between classification accuracy and training effort as was shown by \citet{lee2018training} for full-precision convolutional SNNs without residual connections. Such a hybrid approach would also preclude the need for using the pooled spiking activations of all the convolutional layers for inference, thereby enhancing the scalability of deep ReStoCNets. 

\subsection{Applicability of ReStoCNet for Neuromorphic Hardware Implementations}
Together with research efforts that are geared towards the exploration of bio-plausible SNN algorithms (architectures and learning methodologies), parallel efforts are underway to develop neuromorphic hardware implementations with on-chip intelligence, which can exploit the inherent computational efficiency offered by the SNN algorithms. IBM \textit{TrueNorth} \citep{merolla2014million} and Intel \textit{Loihi} \citep{davies2018loihi} are recent demonstrations of event-driven neuromorphic hardware that were realized using the conventional CMOS technology. CMOS-based neuromorphic hardware implementations are area- and power-intensive because of the mismatch between the spiking neuronal/synaptic circuits and the neuroscience processes governing their dynamics. In this regard, nanoelectronic devices such as Ag-Si memristor \citep{jo2010nanoscale}, Phase-Change Memory (PCM) \citep{suri2011phase}, Resistive Random Access Memory \citep{rajendran2013specifications} and domain-wall Magnetic Tunnel Junctions (MTJs) \citep{sengupta2016hybrid} that are capable of naturally mimicking multilevel synaptic dynamics have been proposed as potential candidates for achieving improved energy efficiency compared to CMOS-only realizations. However, as the technology is scaled, the multilevel memrisitve and spintronic devices suffer from limited bit-precision and exhibit stochastic behavior in the presence of thermal noise. The proposed ReStoCNet, which is composed of binary kernels trained using probabilistic HB-STDP, is naturally suited for neuromorphic hardware implementations based on stochastic device technologies as elaborated in the following paragraph.

Stochastic device technologies such as Conductive-Bridge Random Access Memory (CBRAM) \citep{suri2013bio}, RRAM \citep{kavehei2014highly}, MTJ \citep{vincent2015spin, sengupta2016magnetic, srinivasan2016magnetic}, and PCM \citep{tuma2016stochastic} have been shown to efficiently implement stochastic neuronal and synaptic models. The intrinsic stochastic switching behavior of these devices can be exploited to realize the probabilistic switching of a binary synapse during training without the need for costly random number generators to implement the stochastic operations as illustrated with MTJ-based synapse. An MTJ is composed of two ferromagnetic layers, namely, a pinned layer whose magnetization is fixed and a free layer whose magnetization can be switched, separated by a tunneling oxide barrier. It exhibits two stable conductance states based on the relative orientation of the pinned layer and free layer magnetizations, which can be switched probabilistically by passing charge current through a Heavy Metal (HM) located underneath the MTJ structure. \citet{srinivasan2016magnetic} showed that the MTJ-HM heterostructure, with independent spike-transmission and programming current paths, can efficiently realize a stochastic binary synapse. During training, the MTJ is switched probabilistically based on the time difference between pre- and post-spikes by passing the appropriate current through the HM. During inference, an input pre-spike gets modulated with the trained MTJ conductance to produce resultant current into the post-neuron. \citet{srinivasan2016magnetic} also presented peripheral circuits required to implement an exponential probabilistic-STDP rule, which needs to be modified for realizing the proposed HB-STDP rule. We note that CBRAM, RRAM, and PCM devices can similarly be used to realize a stochastic binary synapse during training by modulating the input voltage based on spike timing \citep{suri2013bio, kavehei2014highly}. Crossbar-based hardware implementations based on these stochastic device technologies with on-chip learning capability have been demonstrated for efficiently realizing binary fully-connected SNNs \citep{suri2013bio, srinivasan2016magnetic}, which consists of a unique synaptic weight connecting every pair of pre- and post-neurons. Recently \citet{wijesinghe2018an} showed that weight-shared convolutional SNNs such as ReStoCNet can be mapped to crossbar-based hardware implementations. However, large-scale networks with increased number of neurons and synapses cannot be mapped to a single large crossbar due to non-idealities that could result in erroneous computations. Hardware architectures composed of multiple smaller crossbars can be used to efficiently realize large-scale networks \citep{shafiee2016isaac, song2017pipelayer, ankit2017resparc}. Finally, we note that the fully-connected classification layer in ReStoCNet, which is composed of artificial ReLU neurons, cannot be directly implemented in event-driven asynchronous neuromorphic hardware platforms. The fully-connected layer of ReLU neurons could be mapped to Integrate-and-Fire neurons post training for inference within the neuromorphic fabric as shown by \citet{diehl2015fast}. Alternatively, fully-connected layer of Leaky-Integrate-and-Fire neurons can be trained using spike-based backpropagation algorithms for training and/or inference within the neuromorphic fabric.

\section{Conclusion}
In this work, we proposed ReStoCNet, a residual stochastic multilayer convolutional SNN composed of binary kernels, for memory-efficient neuromorphic computing. We presented probabilistic Hybrid-STDP (HB-STDP) learning rule, integrating Hebbian and anti-Hebbian learning mechanisms, for training the binary kernels constituting ReStoCNet in a layer-wise unsupervised manner. We demonstrated up to 3-layer deep ReStoCNet and showed that residual connections are critical to enabling the deeper convolutional layers to self-learn useful high-level input features and improving the scalability of deep SNNs. ReStoCNet offered 98.54\% accuracy and 39.5$\times$ kernel memory compression compared to full-precision (32-bit) convolutional SNN under iso-accuracy conditions for MNIST digit recognition. On the CIFAR-10 dataset, ReStoCNet provided 66.23\% accuracy and 21.7$\times$ kernel memory compression, albeit with 5\% accuracy degradation compared to full-precision convolutional SNN. We believe that ReStoCNet, with event-driven computing capability and memory-efficient probabilistic learning with binary kernels, is ideally suited for neuromorphic hardware implementations based on CMOS and stochastic emerging device technologies like Resistive Random Access Memory, Phase-Change Memory, and Magnetic Tunnel Junctions that can potentially lead to much improved energy efficiency in battery-powered IoT edge devices.

\section*{Conflict of Interest Statement}
The authors declare that the research was conducted in the absence of any commercial or financial relationships that could be construed as a potential conflict of interest.

\section*{Author Contributions}
GS wrote the paper and performed the simulations. All authors helped with developing the concepts, conceiving the experiments, and writing the paper.

\section*{Acknowledgments}
This work was supported in part by the Center for Brain Inspired Computing (C-BRIC), one of the six centers in JUMP, a Semiconductor Research Corporation (SRC) program sponsored by DARPA, by the Semiconductor Research Corporation, the National Science Foundation, Intel Corporation, the DoD Vannevar Bush Fellowship, and by the U.S. Army Research Laboratory and the U.K. Ministry of Defense under Agreement Number W911NF-16-3-0001. The views and conclusions contained in this document are those of the authors and should not be interpreted as representing the official policies, either expressed or implied, of the U.S. Army Research Laboratory, the U.S. Government, the U.K. Ministry of Defence or the U.K. Government.

\bibliographystyle{frontiersinSCNS_ENG_HUMS} 
\bibliography{StocConvSNN_ref}

\begin{table}[t]
  \caption{Simulation parameters for training the binary fully-connected SNN on the MNIST dataset.}
  \label{table:fcn_param}
  \resizebox{0.44\textwidth}{!}{%
  \begin{tabular}{ll}
    \hline
    \hline
    Parameters & Values  \\
    \hline
    Simulation time-step, $\Delta t_{sim}$         & 0.5$ms$    \\
    Simulation period, $T_{sim}$                   & 350$ms$    \\
    Maximum input spike rate                       & 63.75$Hz$  \\
    Pre-trace time constant, $\tau_{pre}$          & 20$ms$     \\
    Post-trace time constant, $\tau_{post}$        & 20$ms$     \\
    $pre_{Hebb\_pot}$ \hspace{0.21in}   (eHB-STDP) & 0.85       \\
    $pre_{antiHebb\_dep}$               (eHB-STDP) & 0.10       \\
    $post_{Hebb\_dep}$ \hspace{0.13in}  (eHB-STDP) & 0.80       \\
    $p_{Hebb\_pot}$ \hspace{0.362in}    (eHB-STDP) & 0.08       \\
    $p_{antiHebb\_dep}$ \hspace{0.11in} (eHB-STDP) & 0.06       \\
    $p_{Hebb\_dep}$ \hspace{0.35in}     (eHB-STDP) & 0.005      \\
    Maximum synaptic weight ($w_{high}$)           & 1.0        \\
    Minimum synaptic weight ($w_{low}$)            & 0.0        \\
    \hline
    \hline
  \end{tabular}}
\end{table}

\begin{table}[t]
  \caption{Classification accuracy of binary fully-connected SNNs on the MNIST test set.}
  \label{table:binary_fcn_mnist}
  \resizebox{0.98\textwidth}{!}{%
  \begin{tabular}{llll}
    \hline
    \hline
    Model & \#Excitatory Neurons & Training Methodology & Accuracy (\%)  \\
    \hline
    Binary SNN \citep{querlioz2015bioinspired} & 50  & Probabilistic Rectangular STDP & 60     \\
    Binary SNN \citep{srinivasan2016magnetic}  & 400 & Probabilistic Exponential STDP & 70.15  \\
    Binary SNN (our work) & 400  & Probabilistic eHB-STDP & 79.94  \\
    Binary SNN (our work) & 6400 & Probabilistic eHB-STDP & 92.14  \\
    \hline
    \hline
  \end{tabular}}
\end{table}

\begin{table}[t]
  \caption{Simulation parameters for training the convolutional layers in ReStoCNet.}
  \label{table:ReStoCNet_conv}
  \resizebox{0.90\textwidth}{!}{%
  \begin{tabular}{llll}
    \hline
    \hline
    \multicolumn{3}{r}{Values}    \\
    \cline{2-4}
    Parameters & C1 & C1 & C2/C3  \\
    \hline
    Input dataset                                   & MNIST & CIFAR-10 & CIFAR-10     \\
    Maximum synaptic weight ($w_{high}$)            & $+$1.0 & $+$1.0 & $+$1.0        \\
    Minimum synaptic weight ($w_{low}$)             & $-$1.0 & $-$1.0 & $-$1.0        \\
    Weight initialization constant ($\alpha_{weight\_init}$) & 75 & 30 & 30           \\
    Simulation time-step, $\Delta t_{sim}$          & 1$ms$ & 1$ms$ & 1$ms$           \\
    Simulation period for STDP, $T_{STDP}$          & 25$ms$ & 25$ms$ & 25$ms$        \\
    Maximum input spike rate for STDP               & 200$Hz$ & 200$Hz$ & 500$Hz$     \\
    Dropout probability for STDP, $p_{drop}$        & 0.5 & 0.5 & 0.5                 \\
    $STDP_{stride}$                                 & 5 & 5 & 5                       \\
    Pre-trace decay time constant, $\tau_{pre}$     & 1.45$ms$ & 1.45$ms$ & 1.45$ms$  \\
    $pre_{Hebb\_pot}$ \hspace{0.21in}    (eHB-STDP) & 0.50e-1 & 0.20e-1 & 0.20e-1     \\
    $pre_{antiHebb\_dep}$                (eHB-STDP) & 0.50e-2 & 0.50e-2 & 0.50e-2     \\
    $p_{Hebb\_pot}$ \hspace{0.362in}     (eHB-STDP) & 0.01 & 0.05 & 0.05/25           \\
    $p_{antiHebb\_dep}$ \hspace{0.108in} (eHB-STDP) & 0.01 & 0.01 & 0.01/25           \\
    $p_{Hebb\_dep}$ \hspace{0.345in}     (eHB-STDP) & 0 & 0 & 0                       \\
    $pre_{Hebb\_dep}$ \hspace{0.178in}   (iHB-STDP) & -- & 0.20e-1 & 0.20e-1          \\
    $pre_{antiHebb\_pot}$                (iHB-STDP) & -- & 0.50e-2 & 0.50e-2          \\
    $p_{Hebb\_dep}$ \hspace{0.33in}      (iHB-STDP) & -- & 0.05 & 0.05/25             \\
    $p_{antiHebb\_pot}$ \hspace{0.113in} (iHB-STDP) & -- & 0.01 & 0.01/25             \\
    $p_{Hebb\_pot}$ \hspace{0.348in}     (iHB-STDP) & -- & 0 & 0                      \\         
    Leaky-Integrate-and-Fire (LIF) neuron leak time constant, $\tau_{mem}$   & 9.5$ms$ & 9.5$ms$ & 9.5$ms$ \\
    Rate of increase of LIF neuronal firing threshold, $\beta_{thresh}$ & 6e-4 & 6e-4 & 6e-4 (C2)          \\
                                                                        &      &      & 8e-4 (C3)          \\ 
    Integrate-and-Fire (IF) neuron pooling threshold, $\theta_{pool}$ & 0.80 & 0.80 & 0.80                 \\ 
    Simulation period to estimate spiking activation, $T_{sim}$       & 100$ms$ & 100$ms$ & 100$ms$        \\
    Maximum input spike rate to estimate spiking activation           & 500$Hz$ & 500$Hz$ & 500$Hz$        \\
Low-pass filter time constant to estimate spiking activation, $\tau_{lpf}$ & 99.5$ms$ & 99.5$ms$ & 99.5$ms$\\
    \hline
    \hline
  \end{tabular}}
\end{table}

\begin{table}[t]
  \caption{Simulation parameters for training the fully-connected layer in ReStoCNet.}
  \label{table:ReStoCNet_fcn}
  \resizebox{0.47\textwidth}{!}{%
  \begin{tabular}{lll}
    \hline
    \hline
    \multicolumn{2}{r}{Values}     \\
    \cline{2-3}
    Parameters & MNIST & CIFAR-10  \\
    \hline
    Batch size                   & 256 & 256        \\
    Number of epochs             & 100 & 100        \\
    Learning rate \hspace{-0.03in} (Adam) & 1.5e-3 & 1.0e-4     \\
    betas \hspace{0.52in} (Adam) & (0.9, 0.999) & (0.9, 0.999)  \\
    eps \hspace{0.64in}   (Adam) & 1e-8 & 1e-8      \\
    Weight decay          (Adam) & 0 & 0            \\
    Dropout probability          & 0.5 & 0.5        \\
    \hline
    \hline
  \end{tabular}}
\end{table}

\begin{table}[t]
  \caption{Classification accuracy of SNN models, which use unsupervised training methodology for the hidden/convolutional layers and supervised training algorithm for the output (classification) layer, on the MNIST test set.}
  \label{table:ReStoCNet_mnist}
  \resizebox{\textwidth}{!}{%
  \begin{tabular}{llll}
    \hline
    \hline
    Model & Size & Training Methodology & Accuracy (\%)  \\
    \hline
    FC\_SNN \citep{yousefzadeh2018on}     & 6400FC-10FC               & Probabilistic STDP +     & 95.70  \\ 
                                          &                           & ANN backpropagation      &        \\
    ConvSNN \citep{panda2016unsupervised} & 12C5-2P-64C5-2P-10FC      & SNN backpropagation      & 99.08  \\
    ConvSNN \citep{stromatias2017event}   & 18C7-2P-10FC              & Fixed Gabor kernels +    & 98.20  \\
                                          &                           & ANN backpropagation      &        \\
    ConvSNN \citep{lee2018deep}           & 16C3-16C3-2P-10FC         & STDP                     & 91.10  \\
    ConvSNN \citep{ferre2018unsupervised} & 8C5-2P-16C5-2P-           & STDP +                   & 98.49  \\
                                          & 120FC-60FC-10FC           & ANN backpropagation      &        \\
    ConvSNN \citep{kheradpisheh2018stdp}  & 30C5-2P-100C5-2P-10FC     & STDP +                   & 98.40  \\
                                          &                           & Support Vector Machine   &        \\
    ConvSNN \citep{tavanaei2018training}  & 64C5-2P-1500FC-10FC       & STDP                     & 98.61  \\
    ConvSNN \citep{mozafari2018combining} & 30C5-2P-250C3-3P-200C5-5P & Reward-modulated STDP    & 97.20  \\
    ConvSNN \citep{tavanaei2017multi}     & 32C5-2P-128FC-10FC        & STDP +                   & 98.36  \\
                                          &                           & Support Vector Machine   &        \\
    ReStoCNet (our work)                  & 36C3-2P-128FC-10FC        & Probabilistic eHB-STDP + & 98.54  \\
                                          &                           & ANN backpropagation      &        \\
    \hline
    \hline
  \end{tabular}}
\end{table}

\begin{table}[t]
  \caption{Classification accuracy of SNN models, which use unsupervised training methodology for the hidden/convolutional layers and supervised training algorithm for the output (classification) layer, on the CIFAR-10 test set.}
  \label{table:ReStoCNet_cifar}
  \resizebox{\textwidth}{!}{%
  \begin{tabular}{llll}
    \hline
    \hline
    Model & Size & Training Methodology & Accuracy (\%)  \\
    \hline
    ConvSNN \citep{panda2016unsupervised} & 32C5-2P-32C5-2P-64C4-10FC & SNN backpropagation        & 70.16  \\
    ConvSNN \citep{ferre2018unsupervised} & 64C7-8P-512FC-512FC-10FC  & STDP +                     & 71.20  \\
                                          &                           & ANN backpropagation        &        \\
    ReStoCNet (our work)                  & 256C3-2P-1024FC-10FC      & Probabilistic e/iHB-STDP + & 66.23  \\
                                          &                           & ANN backpropagation        &        \\
    \hline
    \hline
  \end{tabular}}
\end{table}

\end{document}